\pdfoutput=1

\documentclass[11pt]{article}

\usepackage[preprint]{acl}

\usepackage{times}
\usepackage{latexsym}

\usepackage[T1]{fontenc}

\usepackage[utf8]{inputenc}

\usepackage{microtype}

\usepackage{inconsolata}

\usepackage{graphicx}

\usepackage{amsmath}
\usepackage{amssymb}
\usepackage{mathtools}
\usepackage{mlmath}
\usepackage{amsthm}

\usepackage{enumitem}
\usepackage{multirow}
\usepackage{makecell}

\theoremstyle{plain}
\newtheorem{theorem}{Theorem}[section]

\newtheorem{lemma}[theorem]{Lemma}

\theoremstyle{definition}

\theoremstyle{remark}

\newcommand\red[1]{\textcolor{red!0!black}{#1}}
\newcommand\blue[1]{\textcolor{blue!0!black}{#1}}
\newcommand\green[1]{\textcolor{green!0!black}{#1}}

\def\va{{\bm{a}}}
\def\vb{{\bm{b}}}

\def\vp{{\bm{p}}}

\def\vx{{\bm{x}}}
\def\vy{{\bm{y}}}
\def\vz{{\bm{z}}}

\def\mI{{\bm{I}}}

\def\mX{{\bm{X}}}
\def\mY{{\bm{Y}}}
\def\mW{{\bm{W}}}
\def\mZ{{\bm{Z}}}

\usepackage{booktabs}   
\usepackage{array}      
\usepackage{multirow}   
\usepackage{caption}    

\title{Understanding the Language Model to Solve the Symbolic Multi-Step Reasoning Problem from the Perspective of Buffer Mechanism}

\author{
 \textbf{Zhiwei Wang\textsuperscript{1,2}},
 \textbf{Yunji Wang\textsuperscript{1,2}},
 \textbf{Zhongwang Zhang\textsuperscript{1,2}},
 \textbf{Zhangchen Zhou\textsuperscript{1,2}},
\\
 \textbf{Hui Jin\textsuperscript{4}},
 \textbf{Tianyang Hu\textsuperscript{3}},
 \textbf{Jiacheng Sun\textsuperscript{4}},
 \textbf{Zhenguo Li \textsuperscript{4}},
\\
 \textbf{Yaoyu Zhang\textsuperscript{1,2}},
 \textbf{Zhi-Qin John Xu\textsuperscript{1,2,5}}
\\
 \textsuperscript{1}Institute of Natural Sciences, MOE-LSC, Shanghai Jiao Tong University\\
 \textsuperscript{2}School of Mathematical Sciences, Shanghai Jiao Tong University\\
 \textsuperscript{3}The Chinese University of Hong Kong, Shenzhen\\
 \textsuperscript{4}Huawei Noah's Ark Lab\\
 \textsuperscript{5}Key Laboratory of Marine Intelligent Equipment and System, Ministry of Education
\\
 \small{
   \textbf{Correspondence:} \href{mailto:email@domain}{xuzhiqin@sjtu.edu.cn}
 }
}

\begin{document}
\maketitle
\begin{abstract}
Large language models have consistently struggled with complex reasoning tasks, such as mathematical problem-solving. Investigating the internal reasoning mechanisms of these models can help us design better model architectures and training strategies, ultimately enhancing their reasoning capability. In this study, we constructed a symbolic multi-step reasoning task to investigate the information propagation mechanisms in Transformer models when solving the task through direct answering and Chain-of-Thought (CoT) reasoning. We introduced the concept of buffer mechanism: the model stores various information in distinct buffers and selectively extracts it through the query-key matrix. We proposed a random matrix-based algorithm to enhance the model's reasoning ability. This algorithm introduces only 132 trainable parameters, yet leads to significant performance improvements on 7 multi-step reasoning datasets, including PrOntoQA, LogicAsker, and LogicInference. These findings provide new insights into understanding the large language models.
\end{abstract}

\section{Introduction}

    In recent years, LLMs have emerged and demonstrated remarkable capability across various tasks \citep{vaswani2017attention, liu2018generating, devlin2018bert, radford2019language, touvron2023llama, openai2023gpt4, liu2024deepseek, guo2024deepseek, guo2025deepseek}. These models have shown impressive in-context learning abilities \citep{brown2020language, dong2022survey, garg2022can} and have been applied to logical reasoning problems, such as matching top human contestants at the International Mathematical Olympiad (IMO) level \citep{trinh2024solving} and solving math problems \citep{davies2021advancing}. However, even the most advanced models still struggle with complex reasoning tasks. To truly enhance the reasoning capability of LLMs, it is crucial to investigate their intrinsic mechanisms. 
    

    Multi-step reasoning tasks encompass a broad concept, typically referring to the ability of a model to synthesize numerous complex conditions to answer questions. Here, we consider a representative class of syntactic structures within multi-step reasoning tasks: a sentence that includes both the question and sufficient known information to answer it. For example, ``Given: \texttt{[A]$\to$[B]...[B]$\to$[C]...}, Question: 2-step reasoning starting from [A]", where ``\texttt{...}" represents other textual content unrelated to logical reasoning. The answer to this question is ``\texttt{[C]}". When a sentence contains only one logical reasoning step, it is often handled by the so-called induction head in Transformer \citep{brown2020language, olsson2022context}. However, multi-step reasoning is not merely a linear accumulation of multiple induction heads but involves more complex mechanisms. 

    \begin{figure*}[htb]
        \centering
        \includegraphics[width=0.8\textwidth]{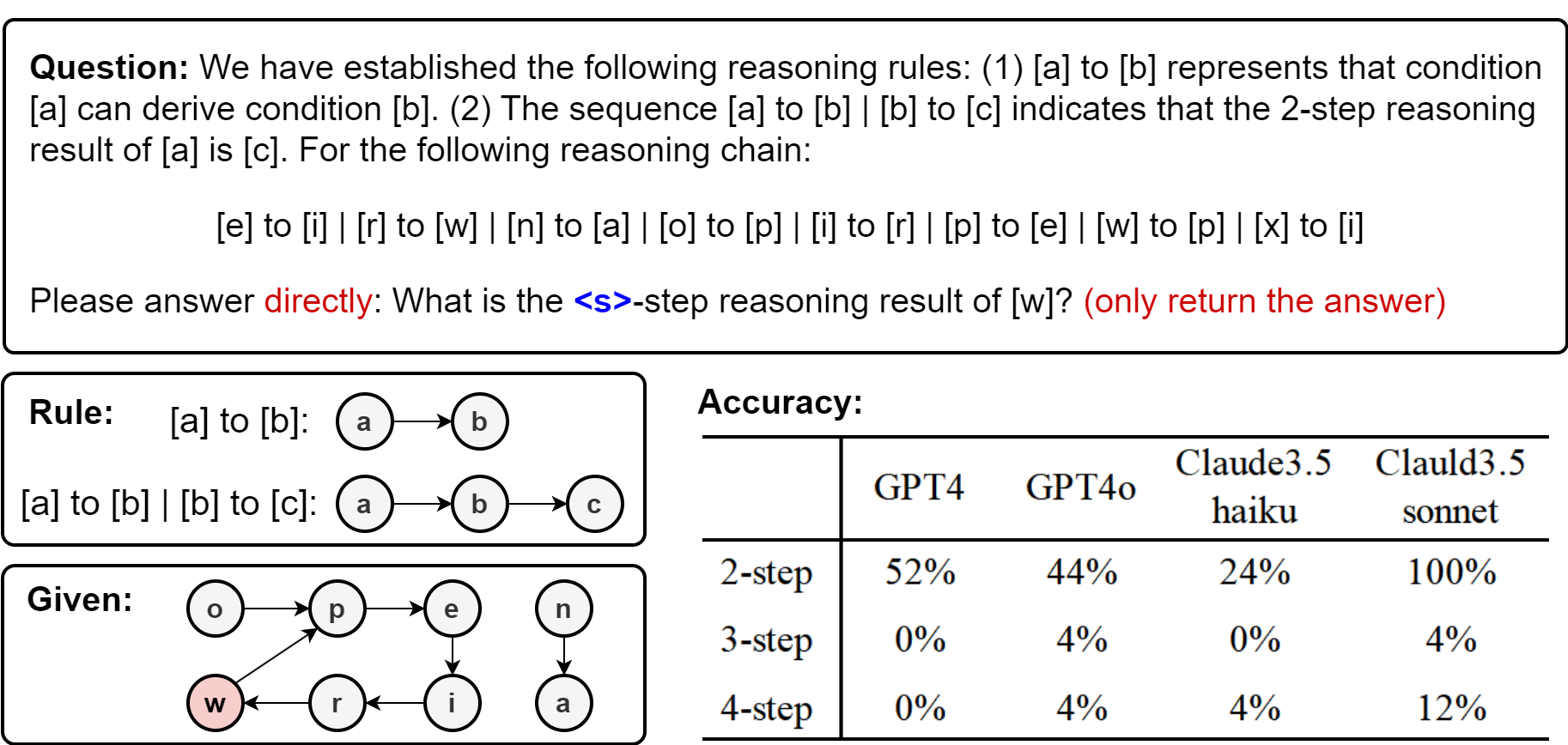}
        \caption{
        \red{The interaction results of multi-step reasoning tasks with large models. We tested each model 25 times, and when the number of reasoning steps exceeded three, these models exhibited random guessing. However, when we allowed the models to use CoT prompting (by removing ``directly" and ``only return the answer"), all models achieved 100\% accuracy. Detailed interaction results are provided in the Appendix~\ref{appendix: QA}.}}
        \label{fig:failure in LLM}
    \end{figure*}

    Large models employ two primary strategies for logical reasoning. The first, known as the \textit{Vertical Thinking Strategy (VTS)}, outputs reasoning results in a single forward based on their inherent structure. This approach is efficient but demands a high level of intelligence. As shown in Fig.~\ref{fig:failure in LLM}, current large models exhibit significant limitations in their vertical thinking capability. Another relatively less efficient approach is the \textit{ (HTS)}, such as Chain of Thought (CoT) \citep{wei2022chain, kojima2022large} and Tree of Thought (ToT) \citep{yao2024tree}, Diagram of Thought (DoT) \citep{zhang2024diagram}. This strategy substantially enhances the model's reasoning performance. All models can produce the correct answers for tasks depicted in Fig.~\ref{fig:failure in LLM} with the help of CoT. Considering the strengths and weaknesses of these two strategies, the ideal approach should combine both: decomposing problems into several coarse-grained subproblems (HTS) and applying the VTS to each subproblem. Thus, researching how to improve vertical thinking capability and understanding why the horizontal thinking strategy can significantly enhance model reasoning ability are both crucial.
    
    In this work, we investigate the performance of Transformer models on a symbolic multi-step reasoning dataset. Our work aims to uncover these mechanisms and provide insights into how Transformers process and integrate logical information across multiple layers to perform multi-step reasoning, which can help develop more effective strategies for improving their multi-step reasoning abilities. Specifically, we found that Transformers utilize a \textbf{\textit{Buffer Mechanism}} when engaging in \red{symbolic multi-step reasoning tasks}. The model stores different intermediate results in separate buffers, allowing for quick retrieval as needed. We elaborate on how the model leverages this buffer mechanism for vertical thinking, and we explain why the horizontal thinking strategy can significantly enhance the model's multi-step reasoning capability from the perspective of the buffer mechanism. Finally, based on this understanding, we propose a method to improve the model's reasoning abilities, leading to significant performance improvements for the GPT-2 model on 7 multi-step reasoning datasets, including Clutrr \citep{sinha2019clutrr}, RuleTaker \citep{clark2020transformers}, StepGame \citep{shi2022stepgame}, LogicInference \citep{ontanon2022logicinference}, LogicAsker \citep{wan2024logicasker}, PararulePlus \citep{bao2022multi}, and PrOntoQA \citep{saparov2022language}.
    
    The concept of ``buffer" or similar concepts has also been mentioned in other works \citep{reddy2023mechanistic, bietti2024birth, elhage2021mathematical}. Our work provides a detailed description of the concept of buffer and applies this mechanism to enhance model performance.

    The main contributions of this work are as follows:
    \begin{itemize}[topsep=0pt, partopsep=0pt, itemsep=0pt, parsep=0pt]
        \item \blue{We propose a buffer mechanism and found evidence that supports such a mechanism being employed by language models during the reasoning process in symbolic multi-step reasoning tasks} and provide a detailed analysis of the model's internal thinking process for vertical and horizontal thinking from the perspective of the buffer.
        \item We propose a method to enhance the model's reasoning capability, improving data utilization efficiency in 7 logical reasoning datasets.
    \end{itemize}
    Our research deepens the understanding of the reasoning mechanisms in Transformer models. 

\section{Reasoning Dataset and Transformer Model}\label{formulation}
    
\begin{figure*}[htb]
        \centering
        \includegraphics[width=0.8\textwidth]{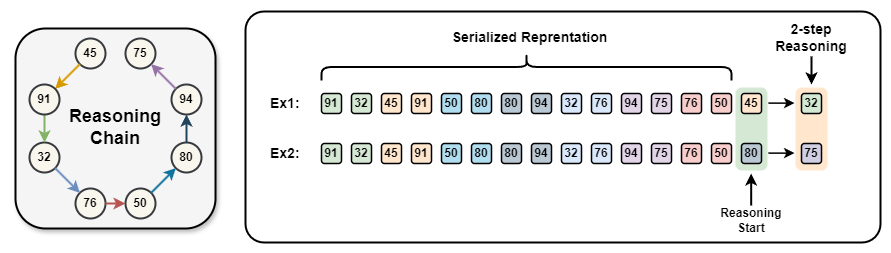}
        \caption{Illustration of the dataset. \red{We investigate reasoning chains composed of digital tokens. In a serialized representation, each pair of adjacent tokens (represented by the same color) forms a reasoning relation within the reasoning chain. The order of the reasoning relations is random, thus a single reasoning chain can correspond to various serialized representations. The model input consists of the serialized representation along with a reasoning start token. The label is the result of performing a fixed number of reasoning steps starting from this start token. This figure illustrates a 2-step reasoning task.}}
        \label{fig:dataset}
    \end{figure*}

    \noindent\textbf{Dataset.} To understand the mechanism of multi-step reasoning in Transformers, we design an abstract symbolic multi-step reasoning task. As shown in Fig.~\ref{fig:dataset}, reasoning chains are serialized into a sequence. Every two tokens in the sentence represent a reasoning relation. The last token is the reasoning start token, and the label is the result with a fixed-step reasoning starting from the starting point. 
        
    \noindent\textbf{Transformer Model.} We employ a decoder-only Transformer. Given an input sequence $\mX^{\mathrm{in}} \in \mathbb{R}^{n \times d}$, where $n$ is the sequence length and $d$ is the dictionary size, the model first applies an embedding layer to obtain the input representation $\mX^{(1)} = \mX_{tgt} + \mX_{pos} \in \mathbb{R}^{n \times d_m}$, where $\mX_{tgt}$ and $\mX_{pos}$ denote the token and positional embeddings, respectively, and $d_m$ is the hidden dimension. The single-head attention in each layer is computed as follows:
    \begin{align*}
        \fA^{(l)}(\mX) &= \sigma\left(\frac{\mathrm{mask}(\mX\mW^{q(l)}\mW^{k(l),\mathsf{T}}\mX^{\mathsf{T}})}{\sqrt{d_k}}\right),\\
        \quad  \mX^{\mathrm{qkv}(l)} &= \fA^{(l)}(\bar{\mX}^{(l)}) \bar{\mX}^{(l)} \mW^{v(l)}\mW^{o(l)},
    \end{align*}
    where $\sigma$ takes the softmax operator, and $\bar{\mX}^{(l)}=\text{Layernorm}(\mX^{(l)})$. For simplicity of expression, we will abbreviate $\mW^{q(l)}\mW^{k(l),\mathsf{T}}$ as $\mW^{qk(l)}$ and $\mW^{v(l)}\mW^{o(l),\mathsf{T}}$ as $\mW^{vo(l)}$ in the following text. The output of the $l$-th layer is obtained as:
    \begin{align*}
        \mX^{\mathrm{ao}(l)} &= \mX^{(l)} + \mX^{\mathrm{qkv}(l)}, \\  \mX^{(l+1)}&=f^{(l)}(\bar{\mX}^{\mathrm{ao}(l)})+\mX^{\mathrm{ao}(l)},
    \end{align*}
    where $f^{(l)}(\cdot)$ represents the feedforward neural network of $l$-th layer. The final output (in the form of token indices within the vocabulary) is obtained as:
    \begin{equation*}
        \mY = argmax(\sigma(\bar{\mX}^{(L)}\mW^{p}))\in \mathbb{R}^{n}.
    \end{equation*}

    \noindent\textbf{In Distribution and Out of Distribution Data.} With the settings of our dataset, if the model truly understands the underlying logic patterns, it should be able to find the correct answer to the sentence, even if this sentence has tokens that have never been encountered during the training. Therefore, we divided the data into two parts: in-distribution (ID) and out-of-distribution (OOD). Specifically, we define $\texttt{token}_\texttt{ID}\in [20,100]$ and $\texttt{token}_\texttt{OOD} \in [0, 120] \setminus [20, 100]$. In-distribution data ($\texttt{Train}_\texttt{ID}$ and $\texttt{Test}_\texttt{ID}$) is defined as sentences composed entirely of $\texttt{token}_\texttt{ID}$, while out-of-distribution data ($\texttt{Test}_\texttt{OOD}$) consists of sentences containing one $\texttt{token}_\texttt{OOD}$, which happens to be the previous reasoning step of label. For the in-distribution data, we split the training set ($\texttt{Train}_\texttt{ID}$) and test set ($\texttt{Test}_\texttt{ID}$) according to the following rules: for the serialized reasoning chain \texttt{[x$_\texttt{1}$][x$_\texttt{2}$]$\cdots$[x$_\texttt{n}$]} of the training set, all tokens satisfy the following condition:
        \begin{equation*}
        \texttt{x}_{\texttt{2i}} - \texttt{x}_{\texttt{2i-1}}\ \ (\text{mod }m) \in G.
        \end{equation*}
    For the reasoning chains in the test set, all tokens satisfy:
        \begin{equation*}
        \texttt{x}_{\texttt{2i}} - \texttt{x}_{\texttt{2i-1}}\ \ (\text{mod }m) \in \{1,\cdots, m\}\backslash G,
        \end{equation*}
    where we take $m=5$ and $G=\{0,1,4\}$ in this study. Under this setting, we ensure that each binary logical pair in the testing set has not previously appeared in the training set. Therefore the Transformer is performing in-context learning \citep{brown2020language, olsson2022context}, as each reasoning pair is not seen during in-weight learning.



    \begin{figure*}[ht]
        \centering
        \includegraphics[width=\textwidth]{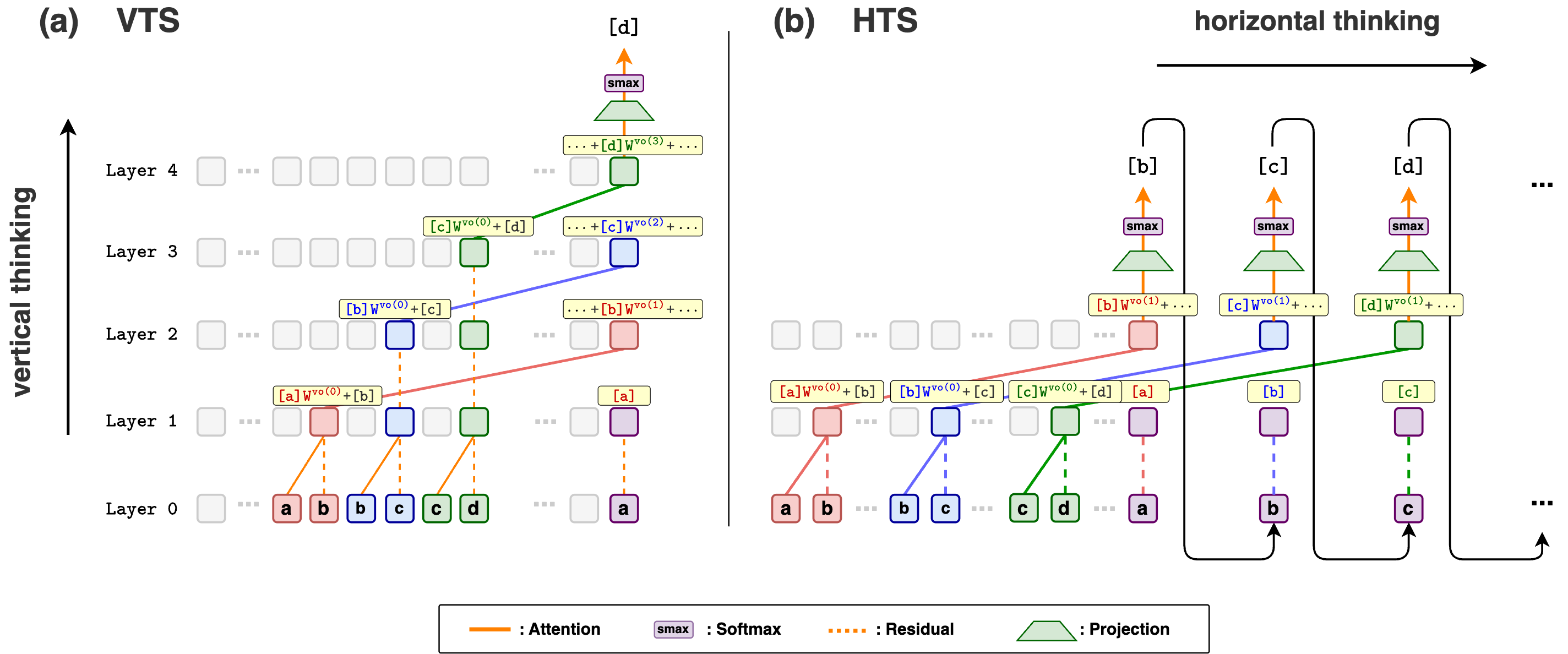}
        \caption{Illustration of how the Transformer employs the vertical and horizontal thinking strategies to solve the multi-step reasoning task. The information crucial for the layer’s information transfer is highlighted. The first attention pairs tokens at odd positions with those at even positions. In the Vertical Thinking Strategy (VTS), each layer conveys its newly derived reasoning result to the final token. In the Horizontal Thinking Strategy (HTS), all information transfer is accomplished within a single layer.}
        \label{fig:VTS_vs_HTS}
    \end{figure*}

\section{Vertical and Horizontal Thinking Strategy}\label{VTS}

    We illustrate how the Transformer model employs vertical or horizontal thinking strategies for multi-step reasoning. Fig.~\ref{fig:VTS_vs_HTS} depicts the schematic mechanisms by which these two strategies execute multi-step inference. The critical logic circuits highlighted in this figure were identified through causal intervention experiments \citep{feng2023language, meng2022locating, vig2020investigating, wang2024grokked} (Appendix~\ref{appendix: causal}). From the figure, one can discern the core distinction between VTS and HTS. Under VTS, in every layer beyond the first—termed a “reasoning layer”—the final token attends to the token carrying the next inference result, conditional on the existing result, thereby effecting a single reasoning step. In contrast, HTS accomplishes all logical inference within a single layer.

    This observation naturally prompts a key question: irrespective of whether VTS or HTS is employed, each node in the Transformer stores multiple pieces of information. For instance, the red node in Layer 1 encodes not only \texttt{[a]} but also \texttt{[b]}. How, then, does the model consolidate diverse information within a single node and accurately extract and utilize the relevant subset? Addressing this question lies at the heart of our investigation into multi-step reasoning.

\section{Buffer Mechanism}

    To address the above problem, we propose a mechanism employed by the Transformer, which we refer to as the “buffer mechanism”. 
    
    Specifically, we observe that the information transmitted via the attention module differs from that conveyed through the residual connections. In the attention module, information \texttt{[a]} with dimension $d_m$ is first mapped to a lower embedding dimension $d_v$ through the matrix $\mW^{v}$, and then it is projected back to $d_m$ dimensions via the matrix $\mW^{o}$. Consequently, the information \texttt{[a]} after passing through the attention layer becomes \texttt{[a]}$\mW^{vo}$. In contrast, information transmitted through residual connections does not require additional processing; thus, after the first layer, the information at even indices transforms to $\texttt{[x}_{\texttt{2i-1}}\texttt{]}\mW^{vo(0)}+\texttt{[x}_{\texttt{2i}}\texttt{]}$.

    Below, we introduce the buffer mechanism within the VTS framework. All theoretical results are supported by experimental validation (Section~\ref{VTS_exp}). We observe that each token in Fig.~\ref{fig:VTS_vs_HTS}(a) stores various information in various forms (projected by different matrices $\mW^{vo(l)}$), which leads to a natural question: how does the attention layer effectively retrieve useful information while avoiding interference from others? To address this question, we introduce the following lemma (the proof can be found in Appendix \ref{proof}):
    \begin{lemma} \label{lemma1}
        Suppose token $\vx=\sum_{i=1}^n \va_i \mW_i \in \mathbb{R}^{d_m}$ and token $\vy=\sum_{i=1}^n \vb_i \mW_i \in \mathbb{R}^{d_m}$, where $\va_i,\vb_i\in \mathbb{R}^{d_m}$, $\mW_i\in \mathbb{R}^{d_m\times d_m}, \ i=1,2,\cdots, n$. Each element of $\{\va_i\}_{i=1}^n$, $\{\vb_i\}_{i=1}^n$ and $\{\mW_i\}_{i=1}^n$ follows $\mathcal{N}(0,1/d_m)$ and independent to others. Then, we have: 
        \begin{align*}
            \vx \mW_i^{\mathsf{T}} &= \va_i + \mathcal{O}\left(\sqrt{\frac{n}{d_m}}\right),\\
            \vy \mW_j^{\mathsf{T}} &= \vb_j + \mathcal{O}\left(\sqrt{\frac{n}{d_m}}\right),
        \end{align*}
        \begin{equation*}
            \vx \mW_i^{\mathsf{T}} \mW_j \vy^{\mathsf{T}} = \va_i \vb_j^{\mathsf{T}} + \mathcal{O}\left(\frac{n}{\sqrt{d_m}}\right).
        \end{equation*}
    \end{lemma}
    It can be observed that the matrices $\{\mW_i\}_{i=1}^n$ serve as a set of \textbf{buffers} for information. Each element of $\{\va_i\}_{i=1}^n$ is located in different buffers, and is almost unaffected by others. This property also applies for $\{\vb_i\}_{i=1}^n$. By selecting the matrices associated with the relevant buffer, specific information contained in token $\vx$ and token $\vy$ can be extracted.
    
    The concept of the buffer mechanism is useful for understanding the internal logic mechanisms of Transformer models. $\{\mW^{q(l)}\}_{l=1}^L$ and $\{\mW^{k(l)}\}_{l=1}^L$ can be viewed as ``information extractors", while $\{\mW^{vo(l)}\}_{l=1}^L$ can be considered as a set of buffers. We observe that in the final token of each reasoning layer, each new intermediate result is always associated with a new $\mW^{vo}$. If these $\{\mW^{vo(l)}\}_{l=1}^L$ matrices are mutually orthogonal or random, then these intermediate results can be regarded as being stored in a new buffer. 
    
    In each reasoning layer, the token at the last position contains the current intermediate result. Additionally, there exists a token that contains both the current intermediate result and the next step's reasoning result, with these stored in different buffers. The role of $\mW^q$ and $\mW^k$ is to extract the current intermediate results from these two tokens, enabling them to attend to each other due to having the same token. We refer to this feature as ``\textit{same-token matching}". To quantitatively characterize this property, we define the following match matrix:
    \begin{align}
        h^{(1)}&(\mX) = \mX\mW^{qk(1)}\mW^{vo(0),\mathsf{T}}\mX^\mathsf{T} \triangleq \mX \, \text{Ker}^{(1)} \, \mX^\mathsf{T},\nonumber\\
        h^{(l)}&(\mX) \\
        &= (\mX\mW^{vo(l-1)})\mW^{q(l)}\mW^{k(l), \mathsf{T}}(\mX\mW^{vo(0)})^\mathsf{T} \nonumber\\
        &=\mX\mW^{vo(l-1)}\mW^{qk(l)}\mW^{vo(0),\mathsf{T}}\mX^\mathsf{T} \nonumber\\
        &\triangleq \mX \, \text{Ker}^{(l)} \, \mX^\mathsf{T},\ l\ge 2.\label{eq:matching_ability_simplified}
    \end{align}
    where 
    \begin{align} 
        \text{Ker}^{(1)}&=\mW^{qk(1)}\mW^{vo(0),\mathsf{T}}, \nonumber\\
        \text{Ker}^{(l)}&=\mW^{vo(l-1)}\mW^{qk(l)}\mW^{vo(0),\mathsf{T}}, \ l\ge 2.\label{eq: ker}
    \end{align} 
    We temporarily ignore the effects introduced by the feedforward layers only in our analysis; a detailed version that includes the feedforward layers can be found in Appendix~\ref{appendix: Matching Operation}. 

    To achieve the same-token matching, it is sufficient for $\text{Ker}^{(l)}\approx I$, in which case 
    \begin{equation}
        h^{(l)}(\mX_{tgt}) \approx \mX_{tgt}\mX_{tgt}^\mathsf{T} = \mI + \mathcal{O}\left(\frac{1}{\sqrt{d_m}}\right).\label{eq: match to I}
    \end{equation}
    Eq.(\ref{eq: match to I}) means that the attention of all tokens is focused on themselves. 
    
    \blue{Furthermore, a much more remarkable observation is that the same-token matching is independent of the specific value of $\mX_{tgt}$.} For example, for $\mX_{tgt,\text{OOD}}$ sampled from the untrained random vectors $\texttt{token}_\texttt{OOD}$, $h^{(l)}(\mX_{tgt,\text{OOD}}) \approx \mI$ still holds. Therefore, when $\text{Ker}^{(l)}\approx I$ holds, the model exhibits OOD generalization capability. 
    
    

    \noindent\textbf{Understanding HTS With Buffer.} Numerous studies have shown that CoT, a canonical representative of HTS, can significantly enhance logical reasoning capability \citep{wei2022chain, kojima2022large}. Fig. \ref{fig:VTS_vs_HTS}(b) presents a schematic diagram illustrating how the model implements this process. In the previously mentioned VTS scenario, the model needed to allocate new buffer $\mW^{vo(1)}$, $\mW^{vo(2)}$ for storing the intermediate result \texttt{[b]} and \texttt{[c]}, respectively. In contrast, with CoT, the model generates a new token to separately store the new intermediate information, effectively replacing the information \texttt{[a]} in the original buffer \red{$\mW^{vo(1)}$}, $\mW^{vo(0)}$ and $\mI$ (identity matrix, which can also be treated as a buffer). Thus, CoT performs multi-step reasoning tasks through buffer reuse. 
    
    \red{Unlike the vertical thinking strategy, which requires alignment across multiple weight matrices, the horizontal thinking strategy only requires a few layers to satisfy $\text{Ker}^{(l)}\approx \mI$. This significantly reduces the difficulty of model training.} \blue{These architectural efficiencies potentially underlie the empirically observed enhancement of reasoning performance in large language models employing CoT methodologies.}

    \color{black}

\section{Experiments}\label{VTS_exp}
    
\begin{figure}[htb]
        \centering
        \includegraphics[width=0.3\textwidth]{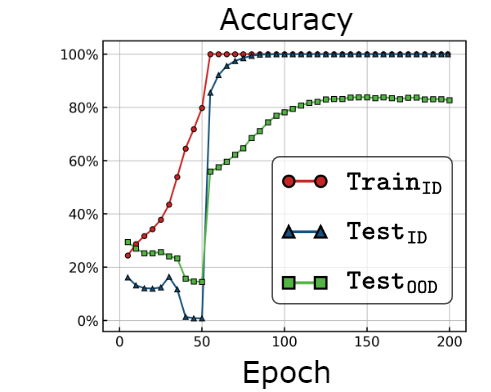}
        \caption{Accuracy curve during VTS training. After training, the Transformer exhibits generalization capability in in-distribution (100\% accuracy) and out-of-distribution data (82\% accuracy).}
        \label{fig:dynamic1}
    \end{figure}

    This section presents the experimental validation of the buffer theory. To demonstrate multi-step reasoning without introducing additional complexity, we utilized a 3-layer, single-head Transformer model to learn from a 2-step reasoning dataset. Specific Experimental settings can be found in Appendix~\ref{appendix: settings}. After training, the Transformer exhibits generalization capability in both in-distribution and out-of-distribution data (Fig.~\ref{fig:dynamic1}).

\begin{figure}[htb]
        \centering
        \includegraphics[width=0.5\textwidth]{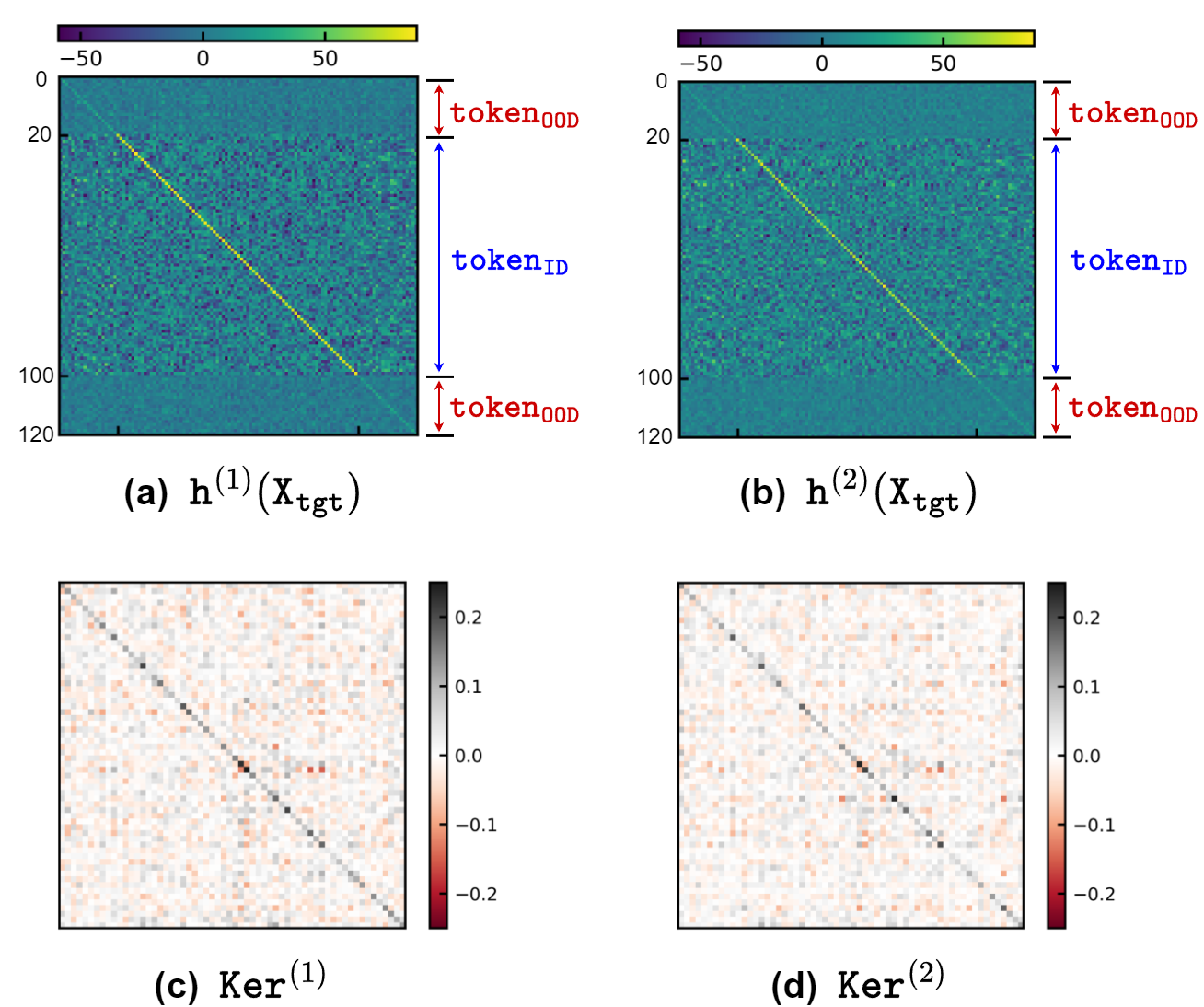}
        \caption{Heatmap of $h^{(1)}(\mX_{tgt})$, $h^{(2)}(\mX_{tgt})$, $\text{Ker}^{(1)}$ and $\text{Ker}^{(2)}$. \red{According to (c)(d), and Eq.(\ref{eq: match to I}), the diagonal structure of the kernel matrix induces a diagonal structure in the matching matrix, even when $\mX_{\text{tgt}}$ is sampled from the $\texttt{token}_\texttt{OOD}$.}}
        \label{fig:matching matrix}
    \end{figure}

    Fig.~\ref{fig:matching matrix} presents the experimental validation of Eq.(\ref{eq: ker})(\ref{eq: match to I}). We visualize the computed values of $h^{(i)}(\mX_{tgt})$ and $\text{Ker}^{(i)}$, $i=1,2$ with the model parameters after training. Notably, in the $\texttt{token}_\texttt{OOD}$ region, $\texttt{token}_\texttt{OOD}$ and $h^{(2)}(\mX_{tgt})$ exhibit a diagonal structure similar to an identity matrix $\mI$, ensuring the model's OOD generalization capability.

    To establish a strong correlation between the model's OOD generalization and its ability to use the buffer mechanism for same-token matching, we define a metric for this capability, the Matching Score (MS):
    \begin{equation}
        \text{MS}(h^{{(l)}}) = \Exp_{\mX}[\text{Trace}(\sigma(h^{(l)}(\mX)))]/n,
    \end{equation}
    where $\mX\in\sR^{n\times d_m}$ is sampled from $\texttt{token}_\texttt{ID}$ or $\texttt{token}_\texttt{OOD}$ for expectation $\Exp_{\mX}$, and $\sigma$ takes the softmax operation. The level of diagonalization of $\text{Ker}^{(l)}$ serves as the intrinsic driver for achieving same-token matching; thus, we also define the following Kernel Score (KS):
    \begin{equation}
        \red{\text{KS}(\text{Ker}^{(l)}) = \text{Trace}(\sigma(\text{Ker}^{(l)}))/d_m}.
    \end{equation}

\begin{figure}[htb]
        \centering
        \includegraphics[width=0.5\textwidth]{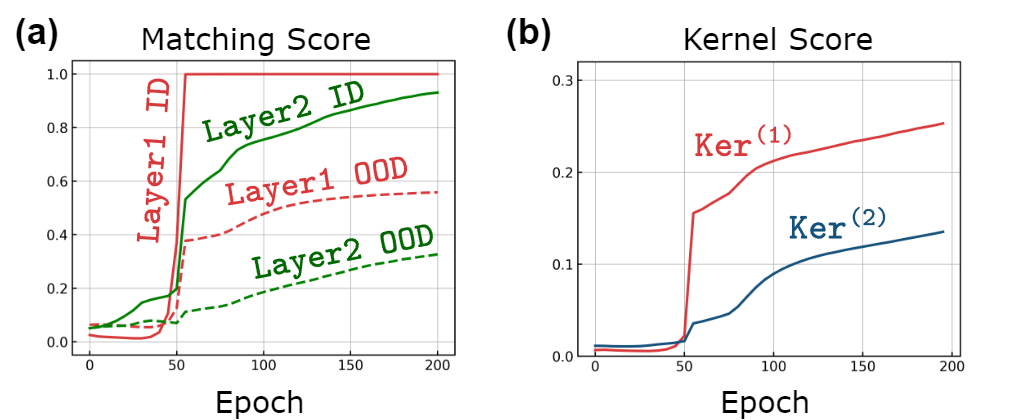}
        \caption{(a) The dynamic evolution of the model's matching score. The red and blue lines represent the matching scores for the first and second layers, respectively. Solid and dashed lines indicate the matching scores when $\mX$ is drawn from $\texttt{token}_\texttt{ID}$ and $\texttt{token}_\texttt{OOD}$, respectively. (b) The kernel scores for the first (red) and second (blue) layers.}
        \label{fig:dynamic2}
    \end{figure}

    Fig.~\ref{fig:dynamic1} and Fig.~\ref{fig:dynamic2} illustrates the accuracy curve and the dynamic changes in the matching score and kernel score during training. It is observed that the increase in the model's ID and OOD generalization coincide with the increases in the model's matching score and kernel score. \blue{By computing the cosine similarity between different buffers ($\mW^{vo(0)},\mW^{vo(1)},\mW^{vo(2)}$ and $I$), we observe that these buffers are nearly pairwise orthogonal(Appendix \ref{appendix: Matching Operation}).} Based on the above experimental analysis, we conclude that the Transformer model indeed employs the buffer mechanism for vertical thinking.

    According to the Buffer mechanism, in theory, by setting the model's weights in the following way, an $L$-layer model can achieve $(L-1)$-step reasoning \textit{without the extra training process}: 
    \begin{equation*}
        \mW^{q(0)}=\mW^{q(1)}=I,\  \mW^{q(l)}=\mW^{vo(l-1),\mathsf{T}},\ l\ge 2, 
    \end{equation*}
    \begin{equation*}
        \mW^{k(0)}=\sum_{i=1}^{[l_\text{seq}/2]}\vp_{2i}\vp_{2i-1}^{\mathsf{T}}, \ \mW^{k(l)}=\mW^{vo(0),\mathsf{T}},\ l\ge 1,
    \end{equation*}
    where $\{\mW^{vo(l)}\}_{l=1}^L$ are set as random matrices and the projection layer satisfies $\mW^{p} = \mW^{vo(L),\mathsf{T}} \mW^{\text{emb},\mathsf{T}}$. Experimental validation can be found in Appendix \ref{appendix: Matching Operation}.

    \noindent\textbf{Real World LLMs.} In Appendix \ref{appendix: Phi-3}, we provide additional definitions for methods to compute the matching score and kernel score in multi-head models and perform these calculations in the real world LLMs such as GPT, Phi, Llama, and Qwen. Similar phenomena, such as weight alignment, provide evidence for the presence of the buffer mechanism in real language models.



\begin{figure}[htb]
        \centering
        \includegraphics[width=0.4\textwidth]{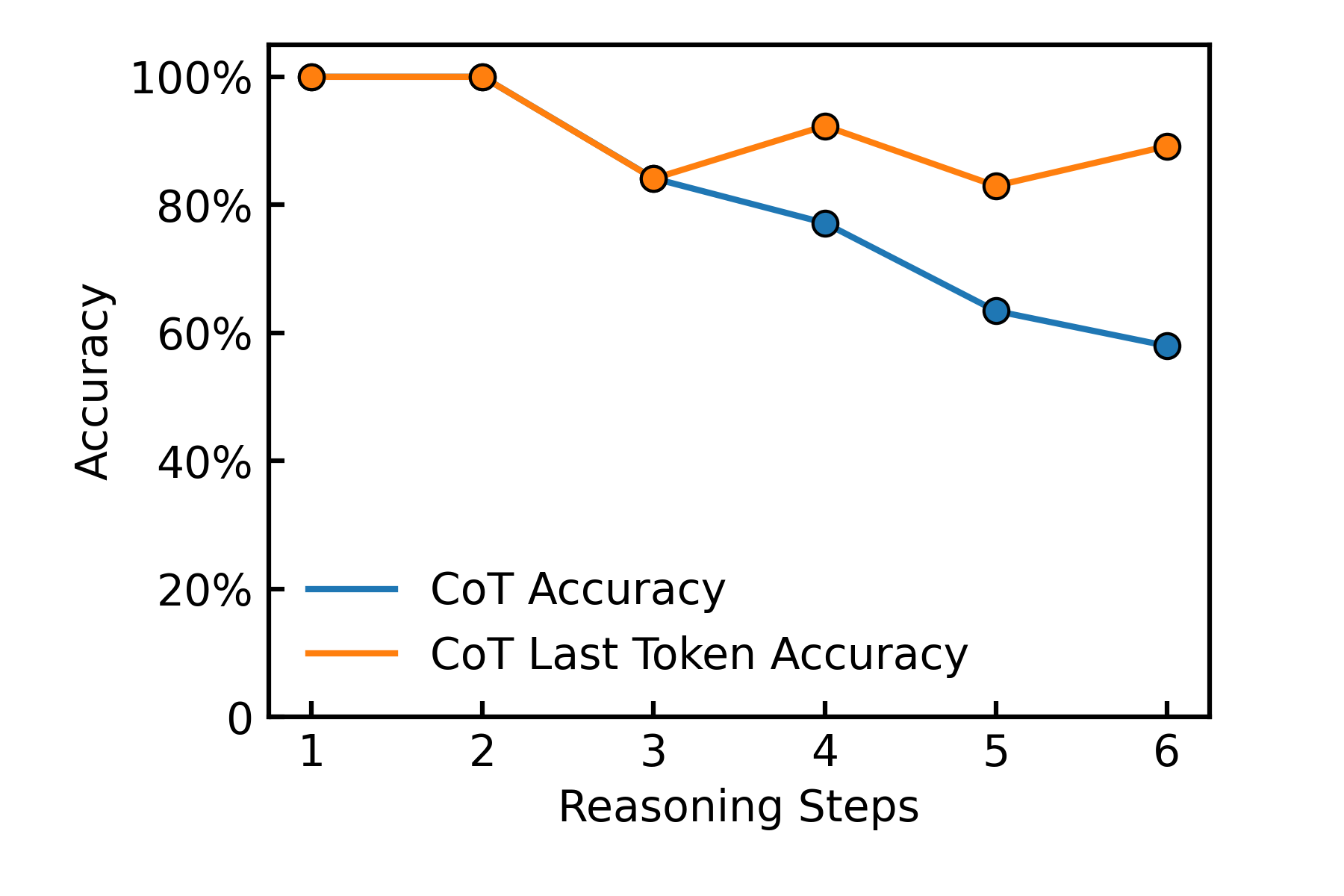}
        \caption{The relationship between the number of reasoning steps and CoT accuracy. The CoT accuracy curve represents the performance when all reasoning steps are predicted correctly using CoT, while the CoT last token accuracy curve records the performance based solely on whether the final reasoning step is correct when the previous correct reasoning result is given.}
        \label{fig:CoT_acc}
    \end{figure}

     \noindent\textbf{Horizontal Thinking Experiments.} For HTS, we will demonstrate that a Transformer model utilizing CoT can achieve arbitrary multi-step reasoning with only 2 layers. We trained a 2-layer Transformer with the \blue{13-length} single-step reasoning data. During the testing phase, we fed the model's output back into the model. Through this CoT process, the model can perform 2-step, 3-step, or even higher-step reasoning \blue{and it can also generalize to sentence lengths beyond the 13th position}. \blue{Fig.~\ref{fig:CoT_acc} shows the relationship between the number of reasoning steps and CoT accuracy. Our 2-layer model is able to maintain an accuracy of over 57.6\% even when performing complex 6-step reasoning tasks.} The specific information flow can be found in Appendix \ref{appendix: CoT}.

\section{Improving Transformer's Data-Efficiency}\label{sec:enhance matching}

    In this section, we discuss how to improve data efficiency when employing the \textbf{vertical thinking strategy}, specifically, by enabling the model to learn the essence of logical reasoning with less data or training epoch. As mentioned in Section~\ref{VTS}, this can be achieved by setting $\mW^{vo(l-1),\mathsf{T}}\mW^{qk(l)}\mW^{vo(0),\mathsf{T}} \approx \mI$.

    A more intuitive approach is to replace $\mW^{qk(l)}$ and $\mW^{vo(l)}$ with $\mW^{qk(l)}+\alpha^{(l)}I$ and $\mW^{vo(l)}+\beta^{(l)}I$, where $\{\alpha^{(l)}\}_{l=1}^L$ and $\{\beta^{(l)}\}_{l=1}^L$ are learnable parameters. This Identity Matrix-Based Algorithm (denoted as IMBA) was first proposed in \citet{boix2023can} and has been validated from both theoretical and empirical perspectives for its role in facilitating the model's learning of one-step reasoning data. However, in multi-step reasoning tasks, IMBA may lead to information interference. For instance, if $\mW^{vo(l)}$ is replaced with $\mW^{vo(l)}+\beta^{(l)}\mI$, the storage representation of the two pieces of information transitions from \texttt{[a]$\mW^{vo}$}+\texttt{[b]} to \texttt{[a]$\mW^{vo}$}+($\beta^{(l)}$\texttt{[a]}+\texttt{[b]}), which introduces interference among the different pieces of information. Therefore, this approach may not be effective in enhancing the model's ability to perform multi-step reasoning.
    
    Based on the understanding of buffer mechanism, we propose a \textbf{Random Matrix-Based Algorithm (RMBA)}, specifically by substituting $\mW^{qk(l)}$ and $\mW^{vo(l)}$ with $\mW^{qk(l)}+\alpha^{(l)}\mZ^{(l-1)}$ and $\mW^{vo(l)}+\beta^{(l)}\mZ^{(l)}$, where $\{\alpha^{(l)}\}_{l=1}^L$ and $\{\beta^{(l)}\}_{l=1}^L$ are learnable parameters and $\{\mZ^{(l)}\}_{l=1}^L$ is a set of fixed random matrix following $N(0,1/d_m)$. In this case, the information storage representation changes from \texttt{[a]$\mW^{vo}$}+\texttt{[b]} to \texttt{[a]$\mW^{vo}$}+\texttt{[a]$\mZ$}+\texttt{[b]}, effectively creating a new buffer.

\begin{figure*}[htb]
        \centering
        \includegraphics[width=0.8\textwidth]{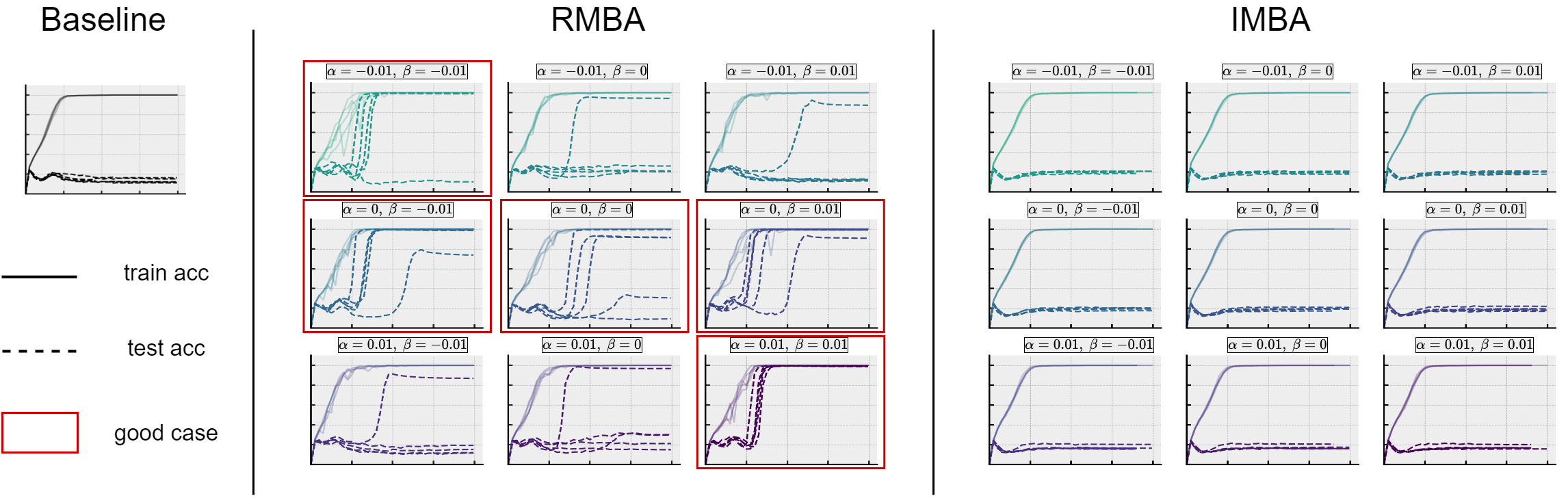}
        \caption{\red{The accuracy comparison for the Baseline model, RMBA model, and IMBA model. The baseline is a three-layer transformer model, which fails to learn a two-step reasoning task with only 30,000 samples. The solid lines represent the training accuracy, while the dashed lines denote the test accuracy. For both RMBA and IMBA models, we set 9 different initialization parameter values $\alpha^{(l)}_{\text{ini}}$ and $\beta^{(l)}_{\text{ini}}$ and each experiment was conducted with 5 random seeds (each seed corresponds to one line).} When $\alpha^{(l)}_{\text{ini}}=0$ or \red{$\alpha^{(l)}_{\text{ini}}\cdot\beta^{(l)}_{\text{ini}}>0$}, the model's reasoning capability can be enhanced. Detailed settings can be found in Appendix \ref{appendix: add noise}.}
        \label{noise id compare}
    \end{figure*}

    In RMBA, when $\alpha^{(l)} \cdot \beta^{(l)} \geq 0$, the diagonal elements of $(\mW^{vo{(l-1)},\mathsf{T}} + \beta^{(l-1)} \mZ^{(l-1),\mathsf{T}})(\mW^{qk{(l)}} + \alpha^{(l)} \mZ^{(l-1)})$ will exhibit positive increments, thereby enhancing the model's reasoning generalization capability. Fig.~\ref{noise id compare} supports this perspective, while also showing that IMBA, based on induction heads, fails to improve multi-step reasoning performance. \red{To strengthen the credibility of our results, we conducted a comprehensive sweep of hyperparameters such as weight decay, learning rate, and hidden dimensions ($d_m$ and $d_k$). \blue{The findings indicate that RMBA parameterization can more robustly reach a high accuracy by a certain number of training steps across a wider range of hyperparameters.} Detailed experimental results can be found in Appendix \ref{appendix: add noise}.}
    
    \blue{These results also demonstrate that multi-step reasoning is not achieved by simply ``stacking" multiple single-step reasonings. An algorithm that enhances single-step reasoning may not be applicable to multi-step reasoning. Therefore, investigating the multi-step reasoning is an important and meaningful topic.}

    \noindent\textbf{Validation of Algorithm Performance on Multiple Real Multi-Step Reasoning Datasets.} In this section, we evaluate the performance of our algorithm using published real-world datasets. The datasets used include Clutrr, RuleTaker, StepGame, LogicInference, LogicAsker, PararulePlus, and PrOntoQA. These datasets assess the model's multi-step reasoning capabilities from different perspectives and scenarios. Examples of the datasets and their corresponding training curves are provided in Appendix~\ref{appendix: add noise}.  

    The 12-layer GPT-2 model is employed for this task. For RMBA, we replace $\mW^{qk(l,h)}$ and $\mW^{vo(l,h)}$ with $\mW^{qk(l,h)} + \alpha^{(l,h)}\mZ^{(l-1,h)}$ and $\mW^{vo(l,h)} + \beta^{(l,h)}\mZ^{(l,h)}$, respectively. To demonstrate the effectiveness of our method, we train each dataset for 5 epochs using a learning rate of $1e\text{-}5$. The additional hyperparameters introduced in RMBA are initialized to 0.05. Training results are summarized in Table~\ref{table1}. For tasks including Clutrr, RuleTalker, StepGame, and LogicInference, RMBA significantly improves model performance. For relatively simpler tasks such as LogicAsker, PararulePlus, and PrOntoQA, although both the base model and RMBA achieve nearly 100\% accuracy, the training curves presented in Fig.~\ref{fig:all_dataset_acc} (Appendix~\ref{appendix: add noise}) show differences in the cost required to reach generalization capability. Table~\ref{table2} reports the number of epochs required to reach 95\% accuracy. As shown, RMBA reduces computational costs by more than 65\% compared to the base model.

    \begin{table}[htbp]
    \centering
    \caption{Final-State Accuracy (\%) Comparison. The training curves for all datasets are shown in Fig.~\ref{fig:all_dataset_acc} (Appendix~\ref{appendix: add noise}).}
    \label{table1}
    \begin{tabular}{lccc}
    \toprule
    Dataset       & Baseline       & \makecell{RMBA\\(ours)}   & IMBA         \\
    \midrule
    Clutrr        & 34.3           & \textbf{42.8}& 42.6 \\
    RuleTaker     & 67.0           & \textbf{83.5}& 74.9 \\
    StepGame      & 35.9           & \textbf{38.9}& 35.3 \\
    LogicInference& 68.1           & \textbf{78.3}& 49.2 \\
    LogicAsker    & 99.7           & 100.0        & 97.1 \\
    PararulePlus  & 99.8           & 99.8         & 99.7 \\
    PrOntoQA      & 100.0          & 100.0        & 100.0\\
    \bottomrule
    \end{tabular}
    \end{table}
    
    \begin{table}[htbp]
    \centering
    \caption{Epochs required to reach 95\% accuracy and cost savings comparison}
    \label{table2}
    \begin{tabular}{lccc}
    \toprule
    Dataset         & Baseline & RMBA(ours)  & IMBA \\
    \midrule
    LogicAsker      & 3.16     & 0.88     & 3.32         \\
    PararulePlus    & 2.71     & 0.73     & 2.02         \\
    PrOntoQA        & 0.95     & 0.32     & 1.25         \\
    \bottomrule
    \end{tabular}
    \end{table}



\section{Discussion}\label{sec: discussion}
    In this study, we investigated the vertical and horizontal thinking strategy employed by Transformer in a symbolic multi-step reasoning task from the perspective of the buffer mechanism. When utilizing the vertical thinking strategy, the model stores different intermediate results in separate buffers and transfers the reasoning results with the same-token matching. In contrast, when applying the horizontal thinking strategy, the model reuses the existing buffers to store intermediate results. We validated that the buffer mechanism is a key factor in enabling the model’s ID and OOD generalization capabilities. Based on the buffer mechanism, we proposed a tailored approach, RMBA, to enhance the model's multi-step reasoning ability, significantly improving data efficiency when training GPT-2 on 7 reasoning datasets.

\section{Limitation}
    Our current work lacks an in-depth theoretical analysis. In future work, we aim to conduct deeper theoretical modeling for multi-step reasoning problems and extend the buffer mechanism to other types of multi-step reasoning tasks.

\bibliography{ref}

\begin{thebibliography}{92}
\providecommand{\natexlab}[1]{#1}

\bibitem[{Abbe et~al.(2022)Abbe, Bengio, Cornacchia, Kleinberg, Lotfi, Raghu, and Zhang}]{abbe2022learning}
Emmanuel Abbe, Samy Bengio, Elisabetta Cornacchia, Jon Kleinberg, Aryo Lotfi, Maithra Raghu, and Chiyuan Zhang. 2022.
\newblock Learning to reason with neural networks: Generalization, unseen data and boolean measures.
\newblock \emph{Advances in Neural Information Processing Systems}, 35:2709--2722.

\bibitem[{Abbe et~al.(2024)Abbe, Bengio, Lotfi, Sandon, and Saremi}]{abbe2024far}
Emmanuel Abbe, Samy Bengio, Aryo Lotfi, Colin Sandon, and Omid Saremi. 2024.
\newblock How far can transformers reason? the locality barrier and inductive scratchpad.
\newblock \emph{arXiv preprint arXiv:2406.06467}.

\bibitem[{Abdin et~al.(2024)Abdin, Jacobs, Awan, Aneja, Awadallah, Awadalla, Bach, Bahree, Bakhtiari, Behl et~al.}]{abdin2024phi}
Marah Abdin, Sam~Ade Jacobs, Ammar~Ahmad Awan, Jyoti Aneja, Ahmed Awadallah, Hany Awadalla, Nguyen Bach, Amit Bahree, Arash Bakhtiari, Harkirat Behl, and 1 others. 2024.
\newblock Phi-3 technical report: A highly capable language model locally on your phone.
\newblock \emph{arXiv preprint arXiv:2404.14219}.

\bibitem[{Amsel et~al.(2024)Amsel, Yehudai, and Bruna}]{amsel2024benefits}
Noah Amsel, Gilad Yehudai, and Joan Bruna. 2024.
\newblock On the benefits of rank in attention layers.
\newblock \emph{arXiv preprint arXiv:2407.16153}.

\bibitem[{Arora et~al.(2018)Arora, Cohen, Golowich, and Hu}]{arora2018convergence}
Sanjeev Arora, Nadav Cohen, Noah Golowich, and Wei Hu. 2018.
\newblock A convergence analysis of gradient descent for deep linear neural networks.
\newblock \emph{arXiv preprint arXiv:1810.02281}.

\bibitem[{Arora et~al.(2019{\natexlab{a}})Arora, Du, Hu, Li, and Wang}]{arora2019fine}
Sanjeev Arora, Simon Du, Wei Hu, Zhiyuan Li, and Ruosong Wang. 2019{\natexlab{a}}.
\newblock Fine-grained analysis of optimization and generalization for overparameterized two-layer neural networks.
\newblock In \emph{International Conference on Machine Learning}, pages 322--332. PMLR.

\bibitem[{Arora et~al.(2019{\natexlab{b}})Arora, Du, Hu, Li, Salakhutdinov, and Wang}]{arora2019exact}
Sanjeev Arora, Simon~S Du, Wei Hu, Zhiyuan Li, Russ~R Salakhutdinov, and Ruosong Wang. 2019{\natexlab{b}}.
\newblock On exact computation with an infinitely wide neural net.
\newblock In \emph{Advances in Neural Information Processing Systems}, pages 8141--8150.

\bibitem[{Arora et~al.(2022)Arora, Li, and Panigrahi}]{arora2022understanding}
Sanjeev Arora, Zhiyuan Li, and Abhishek Panigrahi. 2022.
\newblock Understanding gradient descent on the edge of stability in deep learning.
\newblock In \emph{International Conference on Machine Learning}, pages 948--1024. PMLR.

\bibitem[{Aubry et~al.(2024)Aubry, Meng, Sugolov, and Papyan}]{aubry2024transformer}
Murdock Aubry, Haoming Meng, Anton Sugolov, and Vardan Papyan. 2024.
\newblock Transformer block coupling and its correlation with generalization in llms.
\newblock \emph{arXiv preprint arXiv:2407.07810}.

\bibitem[{Bao et~al.(2022)Bao, Peng, Hartill, Tan, Deng, Witbrock, and Liu}]{bao2022multi}
Qiming Bao, Alex~Yuxuan Peng, Tim Hartill, Neset Tan, Zhenyun Deng, Michael Witbrock, and Jiamou Liu. 2022.
\newblock Multi-step deductive reasoning over natural language: An empirical study on out-of-distribution generalisation.
\newblock \emph{arXiv preprint arXiv:2207.14000}.

\bibitem[{Bietti et~al.(2024)Bietti, Cabannes, Bouchacourt, Jegou, and Bottou}]{bietti2024birth}
Alberto Bietti, Vivien Cabannes, Diane Bouchacourt, Herve Jegou, and Leon Bottou. 2024.
\newblock Birth of a transformer: A memory viewpoint.
\newblock \emph{Advances in Neural Information Processing Systems}, 36.

\bibitem[{Biran et~al.(2024)Biran, Gottesman, Yang, Geva, and Globerson}]{biran2024hopping}
Eden Biran, Daniela Gottesman, Sohee Yang, Mor Geva, and Amir Globerson. 2024.
\newblock Hopping too late: Exploring the limitations of large language models on multi-hop queries.
\newblock \emph{arXiv preprint arXiv:2406.12775}.

\bibitem[{Boix-Adsera et~al.(2023)Boix-Adsera, Saremi, Abbe, Bengio, Littwin, and Susskind}]{boix2023can}
Enric Boix-Adsera, Omid Saremi, Emmanuel Abbe, Samy Bengio, Etai Littwin, and Joshua Susskind. 2023.
\newblock When can transformers reason with abstract symbols?
\newblock \emph{arXiv preprint arXiv:2310.09753}.

\bibitem[{Brinkmann et~al.(2024)Brinkmann, Sheshadri, Levoso, Swoboda, and Bartelt}]{brinkmann2024mechanistic}
Jannik Brinkmann, Abhay Sheshadri, Victor Levoso, Paul Swoboda, and Christian Bartelt. 2024.
\newblock A mechanistic analysis of a transformer trained on a symbolic multi-step reasoning task.
\newblock \emph{arXiv preprint arXiv:2402.11917}.

\bibitem[{Brown et~al.(2020)Brown, Mann, Ryder, Subbiah, Kaplan, Dhariwal, Neelakantan, Shyam, Sastry, Askell et~al.}]{brown2020language}
Tom Brown, Benjamin Mann, Nick Ryder, Melanie Subbiah, Jared~D Kaplan, Prafulla Dhariwal, Arvind Neelakantan, Pranav Shyam, Girish Sastry, Amanda Askell, and 1 others. 2020.
\newblock Language models are few-shot learners.
\newblock \emph{Advances in neural information processing systems}, 33:1877--1901.

\bibitem[{Chen et~al.(2024{\natexlab{a}})Chen, Bruna, and Bietti}]{chen2024truncating}
Lei Chen, Joan Bruna, and Alberto Bietti. 2024{\natexlab{a}}.
\newblock How truncating weights improves reasoning in language models.
\newblock \emph{arXiv preprint arXiv:2406.03068}.

\bibitem[{Chen et~al.(2024{\natexlab{b}})Chen, Sheen, Wang, and Yang}]{chen2024training}
Siyu Chen, Heejune Sheen, Tianhao Wang, and Zhuoran Yang. 2024{\natexlab{b}}.
\newblock Training dynamics of multi-head softmax attention for in-context learning: Emergence, convergence, and optimality.
\newblock \emph{arXiv preprint arXiv:2402.19442}.

\bibitem[{Chen and Zou(2024)}]{chen2024can}
Xingwu Chen and Difan Zou. 2024.
\newblock What can transformer learn with varying depth? case studies on sequence learning tasks.
\newblock \emph{arXiv preprint arXiv:2404.01601}.

\bibitem[{Clark et~al.(2020)Clark, Tafjord, and Richardson}]{clark2020transformers}
Peter Clark, Oyvind Tafjord, and Kyle Richardson. 2020.
\newblock Transformers as soft reasoners over language.
\newblock \emph{arXiv preprint arXiv:2002.05867}.

\bibitem[{Conmy et~al.(2023)Conmy, Mavor-Parker, Lynch, Heimersheim, and Garriga-Alonso}]{conmy2023towards}
Arthur Conmy, Augustine Mavor-Parker, Aengus Lynch, Stefan Heimersheim, and Adri{\`a} Garriga-Alonso. 2023.
\newblock Towards automated circuit discovery for mechanistic interpretability.
\newblock \emph{Advances in Neural Information Processing Systems}, 36:16318--16352.

\bibitem[{Dar et~al.(2022)Dar, Geva, Gupta, and Berant}]{dar2022analyzing}
Guy Dar, Mor Geva, Ankit Gupta, and Jonathan Berant. 2022.
\newblock Analyzing transformers in embedding space.
\newblock \emph{arXiv preprint arXiv:2209.02535}.

\bibitem[{Davies et~al.(2021)Davies, Veli{\v{c}}kovi{\'c}, Buesing, Blackwell, Zheng, Toma{\v{s}}ev, Tanburn, Battaglia, Blundell, Juh{\'a}sz et~al.}]{davies2021advancing}
Alex Davies, Petar Veli{\v{c}}kovi{\'c}, Lars Buesing, Sam Blackwell, Daniel Zheng, Nenad Toma{\v{s}}ev, Richard Tanburn, Peter Battaglia, Charles Blundell, Andr{\'a}s Juh{\'a}sz, and 1 others. 2021.
\newblock Advancing mathematics by guiding human intuition with ai.
\newblock \emph{Nature}, 600(7887):70--74.

\bibitem[{Devlin et~al.(2018)Devlin, Chang, Lee, and Toutanova}]{devlin2018bert}
Jacob Devlin, Ming-Wei Chang, Kenton Lee, and Kristina Toutanova. 2018.
\newblock Bert: Pre-training of deep bidirectional transformers for language understanding.
\newblock \emph{arXiv preprint arXiv:1810.04805}.

\bibitem[{Dong et~al.(2022)Dong, Li, Dai, Zheng, Wu, Chang, Sun, Xu, and Sui}]{dong2022survey}
Qingxiu Dong, Lei Li, Damai Dai, Ce~Zheng, Zhiyong Wu, Baobao Chang, Xu~Sun, Jingjing Xu, and Zhifang Sui. 2022.
\newblock A survey on in-context learning.
\newblock \emph{arXiv preprint arXiv:2301.00234}.

\bibitem[{Dutta et~al.(2024)Dutta, Singh, Chakrabarti, and Chakraborty}]{dutta2024think}
Subhabrata Dutta, Joykirat Singh, Soumen Chakrabarti, and Tanmoy Chakraborty. 2024.
\newblock How to think step-by-step: A mechanistic understanding of chain-of-thought reasoning.
\newblock \emph{arXiv preprint arXiv:2402.18312}.

\bibitem[{Edelman et~al.(2024)Edelman, Edelman, Goel, Malach, and Tsilivis}]{edelman2024evolution}
Benjamin~L Edelman, Ezra Edelman, Surbhi Goel, Eran Malach, and Nikolaos Tsilivis. 2024.
\newblock The evolution of statistical induction heads: In-context learning markov chains.
\newblock \emph{arXiv preprint arXiv:2402.11004}.

\bibitem[{Elhage et~al.(2021)Elhage, Nanda, Olsson, Henighan, Joseph, Mann, Askell, Bai, Chen, Conerly, DasSarma, Drain, Ganguli, Hatfield-Dodds, Hernandez, Jones, Kernion, Lovitt, Ndousse, Amodei, Brown, Clark, Kaplan, McCandlish, and Olah}]{elhage2021mathematical}
Nelson Elhage, Neel Nanda, Catherine Olsson, Tom Henighan, Nicholas Joseph, Ben Mann, Amanda Askell, Yuntao Bai, Anna Chen, Tom Conerly, Nova DasSarma, Dawn Drain, Deep Ganguli, Zac Hatfield-Dodds, Danny Hernandez, Andy Jones, Jackson Kernion, Liane Lovitt, Kamal Ndousse, and 6 others. 2021.
\newblock A mathematical framework for transformer circuits.
\newblock \emph{Transformer Circuits Thread}.
\newblock Https://transformer-circuits.pub/2021/framework/index.html.

\bibitem[{Feng and Steinhardt(2023)}]{feng2023language}
Jiahai Feng and Jacob Steinhardt. 2023.
\newblock How do language models bind entities in context?
\newblock \emph{arXiv preprint arXiv:2310.17191}.

\bibitem[{Garg et~al.(2022)Garg, Tsipras, Liang, and Valiant}]{garg2022can}
Shivam Garg, Dimitris Tsipras, Percy~S Liang, and Gregory Valiant. 2022.
\newblock What can transformers learn in-context? a case study of simple function classes.
\newblock \emph{Advances in Neural Information Processing Systems}, 35:30583--30598.

\bibitem[{Goldowsky-Dill et~al.(2023)Goldowsky-Dill, MacLeod, Sato, and Arora}]{goldowsky2023localizing}
Nicholas Goldowsky-Dill, Chris MacLeod, Lucas Sato, and Aryaman Arora. 2023.
\newblock Localizing model behavior with path patching.
\newblock \emph{arXiv preprint arXiv:2304.05969}.

\bibitem[{Guo et~al.(2025)Guo, Yang, Zhang, Song, Zhang, Xu, Zhu, Ma, Wang, Bi et~al.}]{guo2025deepseek}
Daya Guo, Dejian Yang, Haowei Zhang, Junxiao Song, Ruoyu Zhang, Runxin Xu, Qihao Zhu, Shirong Ma, Peiyi Wang, Xiao Bi, and 1 others. 2025.
\newblock Deepseek-r1: Incentivizing reasoning capability in llms via reinforcement learning.
\newblock \emph{arXiv preprint arXiv:2501.12948}.

\bibitem[{Guo et~al.(2024)Guo, Zhu, Yang, Xie, Dong, Zhang, Chen, Bi, Wu, Li et~al.}]{guo2024deepseek}
Daya Guo, Qihao Zhu, Dejian Yang, Zhenda Xie, Kai Dong, Wentao Zhang, Guanting Chen, Xiao Bi, Yu~Wu, YK~Li, and 1 others. 2024.
\newblock Deepseek-coder: When the large language model meets programming--the rise of code intelligence.
\newblock \emph{arXiv preprint arXiv:2401.14196}.

\bibitem[{Guo et~al.(2023)Guo, Hu, Mei, Wang, Xiong, Savarese, and Bai}]{guo2023transformers}
Tianyu Guo, Wei Hu, Song Mei, Huan Wang, Caiming Xiong, Silvio Savarese, and Yu~Bai. 2023.
\newblock How do transformers learn in-context beyond simple functions? a case study on learning with representations.
\newblock \emph{arXiv preprint arXiv:2310.10616}.

\bibitem[{Jacot et~al.(2018)Jacot, Gabriel, and Hongler}]{jacot2018neural}
Arthur Jacot, Franck Gabriel, and Cl{\'e}ment Hongler. 2018.
\newblock Neural tangent kernel: Convergence and generalization in neural networks.
\newblock \emph{Advances in neural information processing systems}, 31.

\bibitem[{Jacot et~al.(2020)Jacot, Simsek, Spadaro, Hongler, and Gabriel}]{jacot2020implicit}
Arthur Jacot, Berfin Simsek, Francesco Spadaro, Cl{\'e}ment Hongler, and Franck Gabriel. 2020.
\newblock Implicit regularization of random feature models.
\newblock In \emph{International Conference on Machine Learning}, pages 4631--4640. PMLR.

\bibitem[{Jeoung and Diesner(2022)}]{jeoung2022changed}
Sullam Jeoung and Jana Diesner. 2022.
\newblock What changed? investigating debiasing methods using causal mediation analysis.
\newblock \emph{arXiv preprint arXiv:2206.00701}.

\bibitem[{Jiang et~al.(2024)Jiang, Xie, Hao, Wang, Mallick, Su, Taylor, and Roth}]{jiang2024peek}
Bowen Jiang, Yangxinyu Xie, Zhuoqun Hao, Xiaomeng Wang, Tanwi Mallick, Weijie~J Su, Camillo~J Taylor, and Dan Roth. 2024.
\newblock A peek into token bias: Large language models are not yet genuine reasoners.
\newblock \emph{arXiv preprint arXiv:2406.11050}.

\bibitem[{Kil et~al.(2024)Kil, Tavazoee, Kang, and Kim}]{kil2024ii}
Jihyung Kil, Farideh Tavazoee, Dongyeop Kang, and Joo-Kyung Kim. 2024.
\newblock Ii-mmr: Identifying and improving multi-modal multi-hop reasoning in visual question answering.
\newblock \emph{arXiv preprint arXiv:2402.11058}.

\bibitem[{Kobayashi et~al.(2020)Kobayashi, Kuribayashi, Yokoi, and Inui}]{kobayashi2020attention}
Goro Kobayashi, Tatsuki Kuribayashi, Sho Yokoi, and Kentaro Inui. 2020.
\newblock Attention is not only a weight: Analyzing transformers with vector norms.
\newblock \emph{arXiv preprint arXiv:2004.10102}.

\bibitem[{Kojima et~al.(2022)Kojima, Gu, Reid, Matsuo, and Iwasawa}]{kojima2022large}
Takeshi Kojima, Shixiang~Shane Gu, Machel Reid, Yutaka Matsuo, and Yusuke Iwasawa. 2022.
\newblock Large language models are zero-shot reasoners.
\newblock \emph{Advances in neural information processing systems}, 35:22199--22213.

\bibitem[{Kovaleva et~al.(2019)Kovaleva, Romanov, Rogers, and Rumshisky}]{kovaleva2019revealing}
Olga Kovaleva, Alexey Romanov, Anna Rogers, and Anna Rumshisky. 2019.
\newblock Revealing the dark secrets of bert.
\newblock \emph{arXiv preprint arXiv:1908.08593}.

\bibitem[{Li et~al.(2024{\natexlab{a}})Li, Liang, Lyu, and Wang}]{li2024making}
Yanyang Li, Shuo Liang, Michael~R Lyu, and Liwei Wang. 2024{\natexlab{a}}.
\newblock Making long-context language models better multi-hop reasoners.
\newblock \emph{arXiv preprint arXiv:2408.03246}.

\bibitem[{Li et~al.(2024{\natexlab{b}})Li, Jiang, Xie, Song, Lian, and Wei}]{li2024understanding}
Zhaoyi Li, Gangwei Jiang, Hong Xie, Linqi Song, Defu Lian, and Ying Wei. 2024{\natexlab{b}}.
\newblock Understanding and patching compositional reasoning in llms.
\newblock \emph{arXiv preprint arXiv:2402.14328}.

\bibitem[{Li et~al.(2024{\natexlab{c}})Li, Liu, Zhou, and Ma}]{li2024chain}
Zhiyuan Li, Hong Liu, Denny Zhou, and Tengyu Ma. 2024{\natexlab{c}}.
\newblock Chain of thought empowers transformers to solve inherently serial problems.
\newblock \emph{arXiv preprint arXiv:2402.12875}.

\bibitem[{Li et~al.(2021)Li, Malladi, and Arora}]{li2021validity}
Zhiyuan Li, Sadhika Malladi, and Sanjeev Arora. 2021.
\newblock On the validity of modeling sgd with stochastic differential equations (sdes).
\newblock \emph{Advances in Neural Information Processing Systems}, 34:12712--12725.

\bibitem[{Liu et~al.(2024)Liu, Feng, Xue, Wang, Wu, Lu, Zhao, Deng, Zhang, Ruan et~al.}]{liu2024deepseek}
Aixin Liu, Bei Feng, Bing Xue, Bingxuan Wang, Bochao Wu, Chengda Lu, Chenggang Zhao, Chengqi Deng, Chenyu Zhang, Chong Ruan, and 1 others. 2024.
\newblock Deepseek-v3 technical report.
\newblock \emph{arXiv preprint arXiv:2412.19437}.

\bibitem[{Liu et~al.(2018)Liu, Saleh, Pot, Goodrich, Sepassi, Kaiser, and Shazeer}]{liu2018generating}
Peter~J Liu, Mohammad Saleh, Etienne Pot, Ben Goodrich, Ryan Sepassi, Lukasz Kaiser, and Noam Shazeer. 2018.
\newblock Generating wikipedia by summarizing long sequences.
\newblock \emph{arXiv preprint arXiv:1801.10198}.

\bibitem[{Luo et~al.(2021)Luo, Xu, Ma, and Zhang}]{luo2021phase}
Tao Luo, Zhi-Qin~John Xu, Zheng Ma, and Yaoyu Zhang. 2021.
\newblock Phase diagram for two-layer relu neural networks at infinite-width limit.
\newblock \emph{Journal of Machine Learning Research}, 22(71):1--47.

\bibitem[{Meng et~al.(2022)Meng, Bau, Andonian, and Belinkov}]{meng2022locating}
Kevin Meng, David Bau, Alex Andonian, and Yonatan Belinkov. 2022.
\newblock Locating and editing factual associations in gpt.
\newblock \emph{Advances in Neural Information Processing Systems}, 35:17359--17372.

\bibitem[{Merullo et~al.(2023)Merullo, Eickhoff, and Pavlick}]{merullo2023circuit}
Jack Merullo, Carsten Eickhoff, and Ellie Pavlick. 2023.
\newblock Circuit component reuse across tasks in transformer language models.
\newblock \emph{arXiv preprint arXiv:2310.08744}.

\bibitem[{M{\"u}ller et~al.(2021)M{\"u}ller, Hollmann, Arango, Grabocka, and Hutter}]{muller2021transformers}
Samuel M{\"u}ller, Noah Hollmann, Sebastian~Pineda Arango, Josif Grabocka, and Frank Hutter. 2021.
\newblock Transformers can do bayesian inference.
\newblock \emph{arXiv preprint arXiv:2112.10510}.

\bibitem[{Nichani et~al.(2024)Nichani, Damian, and Lee}]{nichani2024transformers}
Eshaan Nichani, Alex Damian, and Jason~D Lee. 2024.
\newblock How transformers learn causal structure with gradient descent.
\newblock \emph{arXiv preprint arXiv:2402.14735}.

\bibitem[{Olsson et~al.(2022)Olsson, Elhage, Nanda, Joseph, DasSarma, Henighan, Mann, Askell, Bai, Chen et~al.}]{olsson2022context}
Catherine Olsson, Nelson Elhage, Neel Nanda, Nicholas Joseph, Nova DasSarma, Tom Henighan, Ben Mann, Amanda Askell, Yuntao Bai, Anna Chen, and 1 others. 2022.
\newblock In-context learning and induction heads.
\newblock \emph{arXiv preprint arXiv:2209.11895}.

\bibitem[{Ontanon et~al.(2022)Ontanon, Ainslie, Cvicek, and Fisher}]{ontanon2022logicinference}
Santiago Ontanon, Joshua Ainslie, Vaclav Cvicek, and Zachary Fisher. 2022.
\newblock Logicinference: A new dataset for teaching logical inference to seq2seq models.
\newblock \emph{arXiv preprint arXiv:2203.15099}.

\bibitem[{OpenAI(2023)}]{openai2023gpt4}
OpenAI. 2023.
\newblock \href {https://arxiv.org/abs/2303.08774} {Gpt-4 technical report}.
\newblock \emph{Preprint}, arXiv:2303.08774.

\bibitem[{Poli et~al.(2024)Poli, Thomas, Nguyen, Ponnusamy, Deiseroth, Kersting, Suzuki, Hie, Ermon, R{\'e} et~al.}]{poli2024mechanistic}
Michael Poli, Armin~W Thomas, Eric Nguyen, Pragaash Ponnusamy, Bj{\"o}rn Deiseroth, Kristian Kersting, Taiji Suzuki, Brian Hie, Stefano Ermon, Christopher R{\'e}, and 1 others. 2024.
\newblock Mechanistic design and scaling of hybrid architectures.
\newblock \emph{arXiv preprint arXiv:2403.17844}.

\bibitem[{Radford et~al.(2019)Radford, Wu, Child, Luan, Amodei, Sutskever et~al.}]{radford2019language}
Alec Radford, Jeffrey Wu, Rewon Child, David Luan, Dario Amodei, Ilya Sutskever, and 1 others. 2019.
\newblock Language models are unsupervised multitask learners.
\newblock \emph{OpenAI blog}, 1(8):9.

\bibitem[{Reddy(2023)}]{reddy2023mechanistic}
Gautam Reddy. 2023.
\newblock The mechanistic basis of data dependence and abrupt learning in an in-context classification task.
\newblock \emph{arXiv preprint arXiv:2312.03002}.

\bibitem[{Ren et~al.(2024)Ren, Ma, and Ying}]{ren2024understanding}
Yinuo Ren, Chao Ma, and Lexing Ying. 2024.
\newblock Understanding the generalization benefits of late learning rate decay.
\newblock In \emph{International Conference on Artificial Intelligence and Statistics}, pages 4465--4473. PMLR.

\bibitem[{Saparov and He(2022)}]{saparov2022language}
Abulhair Saparov and He~He. 2022.
\newblock Language models are greedy reasoners: A systematic formal analysis of chain-of-thought.
\newblock \emph{arXiv preprint arXiv:2210.01240}.

\bibitem[{Shalev et~al.(2024)Shalev, Feder, and Goldstein}]{shalev2024distributional}
Yuval Shalev, Amir Feder, and Ariel Goldstein. 2024.
\newblock Distributional reasoning in llms: Parallel reasoning processes in multi-hop reasoning.
\newblock \emph{arXiv preprint arXiv:2406.13858}.

\bibitem[{Sharma et~al.(2023)Sharma, Ash, and Misra}]{sharma2023truth}
Pratyusha Sharma, Jordan~T Ash, and Dipendra Misra. 2023.
\newblock The truth is in there: Improving reasoning in language models with layer-selective rank reduction.
\newblock \emph{arXiv preprint arXiv:2312.13558}.

\bibitem[{Shi et~al.(2022)Shi, Zhang, and Lipani}]{shi2022stepgame}
Zhengxiang Shi, Qiang Zhang, and Aldo Lipani. 2022.
\newblock Stepgame: A new benchmark for robust multi-hop spatial reasoning in texts.
\newblock In \emph{Proceedings of the AAAI conference on artificial intelligence}, 10, pages 11321--11329.

\bibitem[{Sinha et~al.(2019)Sinha, Sodhani, Dong, Pineau, and Hamilton}]{sinha2019clutrr}
Koustuv Sinha, Shagun Sodhani, Jin Dong, Joelle Pineau, and William~L Hamilton. 2019.
\newblock Clutrr: A diagnostic benchmark for inductive reasoning from text.
\newblock \emph{arXiv preprint arXiv:1908.06177}.

\bibitem[{Todd et~al.(2023)Todd, Li, Sharma, Mueller, Wallace, and Bau}]{todd2023function}
Eric Todd, Millicent~L Li, Arnab~Sen Sharma, Aaron Mueller, Byron~C Wallace, and David Bau. 2023.
\newblock Function vectors in large language models.
\newblock \emph{arXiv preprint arXiv:2310.15213}.

\bibitem[{Touvron et~al.(2023)Touvron, Lavril, Izacard, Martinet, Lachaux, Lacroix, Rozi{\`e}re, Goyal, Hambro, Azhar et~al.}]{touvron2023llama}
Hugo Touvron, Thibaut Lavril, Gautier Izacard, Xavier Martinet, Marie-Anne Lachaux, Timoth{\'e}e Lacroix, Baptiste Rozi{\`e}re, Naman Goyal, Eric Hambro, Faisal Azhar, and 1 others. 2023.
\newblock Llama: Open and efficient foundation language models.
\newblock \emph{arXiv preprint arXiv:2302.13971}.

\bibitem[{Trinh et~al.(2024)Trinh, Wu, Le, He, and Luong}]{trinh2024solving}
Trieu~H Trinh, Yuhuai Wu, Quoc~V Le, He~He, and Thang Luong. 2024.
\newblock Solving olympiad geometry without human demonstrations.
\newblock \emph{Nature}, 625(7995):476--482.

\bibitem[{Vaswani et~al.(2017)Vaswani, Shazeer, Parmar, Uszkoreit, Jones, Gomez, Kaiser, and Polosukhin}]{vaswani2017attention}
Ashish Vaswani, Noam Shazeer, Niki Parmar, Jakob Uszkoreit, Llion Jones, Aidan~N Gomez, {\L}ukasz Kaiser, and Illia Polosukhin. 2017.
\newblock Attention is all you need.
\newblock \emph{Advances in neural information processing systems}, 30.

\bibitem[{Vig(2019)}]{vig2019multiscale}
Jesse Vig. 2019.
\newblock A multiscale visualization of attention in the transformer model.
\newblock \emph{arXiv preprint arXiv:1906.05714}.

\bibitem[{Vig et~al.(2020)Vig, Gehrmann, Belinkov, Qian, Nevo, Singer, and Shieber}]{vig2020investigating}
Jesse Vig, Sebastian Gehrmann, Yonatan Belinkov, Sharon Qian, Daniel Nevo, Yaron Singer, and Stuart Shieber. 2020.
\newblock Investigating gender bias in language models using causal mediation analysis.
\newblock \emph{Advances in neural information processing systems}, 33:12388--12401.

\bibitem[{Voita et~al.(2019)Voita, Talbot, Moiseev, Sennrich, and Titov}]{voita2019analyzing}
Elena Voita, David Talbot, Fedor Moiseev, Rico Sennrich, and Ivan Titov. 2019.
\newblock Analyzing multi-head self-attention: Specialized heads do the heavy lifting, the rest can be pruned.
\newblock \emph{arXiv preprint arXiv:1905.09418}.

\bibitem[{Wan et~al.(2024)Wan, Wang, Yang, Yuan, Huang, He, Jiao, and Lyu}]{wan2024logicasker}
Yuxuan Wan, Wenxuan Wang, Yiliu Yang, Youliang Yuan, Jen-tse Huang, Pinjia He, Wenxiang Jiao, and Michael~R Lyu. 2024.
\newblock Logicasker: Evaluating and improving the logical reasoning ability of large language models.
\newblock \emph{arXiv preprint arXiv:2401.00757}.

\bibitem[{Wang et~al.(2024{\natexlab{a}})Wang, Yue, Su, and Sun}]{wang2024grokked}
Boshi Wang, Xiang Yue, Yu~Su, and Huan Sun. 2024{\natexlab{a}}.
\newblock Grokked transformers are implicit reasoners: A mechanistic journey to the edge of generalization.
\newblock \emph{arXiv preprint arXiv:2405.15071}.

\bibitem[{Wang et~al.(2022)Wang, Variengien, Conmy, Shlegeris, and Steinhardt}]{wang2022interpretability}
Kevin Wang, Alexandre Variengien, Arthur Conmy, Buck Shlegeris, and Jacob Steinhardt. 2022.
\newblock Interpretability in the wild: a circuit for indirect object identification in gpt-2 small.
\newblock \emph{arXiv preprint arXiv:2211.00593}.

\bibitem[{Wang et~al.(2023)Wang, Li, Dai, Chen, Zhou, Meng, Zhou, and Sun}]{wang2023label}
Lean Wang, Lei Li, Damai Dai, Deli Chen, Hao Zhou, Fandong Meng, Jie Zhou, and Xu~Sun. 2023.
\newblock Label words are anchors: An information flow perspective for understanding in-context learning.
\newblock \emph{arXiv preprint arXiv:2305.14160}.

\bibitem[{Wang et~al.(2024{\natexlab{b}})Wang, He, Wang, Wang, Huang, Xiong, Li, Wu et~al.}]{wang2024improving}
Mingze Wang, Haotian He, Jinbo Wang, Zilin Wang, Guanhua Huang, Feiyu Xiong, Zhiyu Li, Lei Wu, and 1 others. 2024{\natexlab{b}}.
\newblock Improving generalization and convergence by enhancing implicit regularization.
\newblock \emph{arXiv preprint arXiv:2405.20763}.

\bibitem[{Wang and Weinan(2024)}]{wang2024understanding}
Mingze Wang and E~Weinan. 2024.
\newblock Understanding the expressive power and mechanisms of transformer for sequence modeling.
\newblock \emph{arXiv preprint arXiv:2402.00522}.

\bibitem[{Wang and Wu(2023)}]{wang2310theoretical}
Mingze Wang and Lei Wu. 2023.
\newblock A theoretical analysis of noise geometry in stochastic gradient descent.
\newblock \emph{arXiv preprint arXiv:2310.00692}.

\bibitem[{Wang et~al.(2024{\natexlab{c}})Wang, Amayuelas, Zhang, Pan, Chen, and Wang}]{wang2024understanding1}
Xinyi Wang, Alfonso Amayuelas, Kexun Zhang, Liangming Pan, Wenhu Chen, and William~Yang Wang. 2024{\natexlab{c}}.
\newblock Understanding the reasoning ability of language models from the perspective of reasoning paths aggregation.
\newblock \emph{arXiv preprint arXiv:2402.03268}.

\bibitem[{Wei et~al.(2022)Wei, Wang, Schuurmans, Bosma, Xia, Chi, Le, Zhou et~al.}]{wei2022chain}
Jason Wei, Xuezhi Wang, Dale Schuurmans, Maarten Bosma, Fei Xia, Ed~Chi, Quoc~V Le, Denny Zhou, and 1 others. 2022.
\newblock Chain-of-thought prompting elicits reasoning in large language models.
\newblock \emph{Advances in neural information processing systems}, 35:24824--24837.

\bibitem[{Wu et~al.(2018)Wu, Ma et~al.}]{wu2018sgd}
Lei Wu, Chao Ma, and 1 others. 2018.
\newblock How sgd selects the global minima in over-parameterized learning: A dynamical stability perspective.
\newblock \emph{Advances in Neural Information Processing Systems}, 31.

\bibitem[{Wu and Su(2023)}]{wu2023implicit}
Lei Wu and Weijie~J Su. 2023.
\newblock The implicit regularization of dynamical stability in stochastic gradient descent.
\newblock In \emph{International Conference on Machine Learning}, pages 37656--37684. PMLR.

\bibitem[{Xu et~al.(2019)Xu, Zhang, Luo, Xiao, and Ma}]{xu2019frequency}
Zhi-Qin~John Xu, Yaoyu Zhang, Tao Luo, Yanyang Xiao, and Zheng Ma. 2019.
\newblock Frequency principle: Fourier analysis sheds light on deep neural networks.
\newblock \emph{arXiv preprint arXiv:1901.06523}.

\bibitem[{Yang et~al.(2024)Yang, Gribovskaya, Kassner, Geva, and Riedel}]{yang2024large}
Sohee Yang, Elena Gribovskaya, Nora Kassner, Mor Geva, and Sebastian Riedel. 2024.
\newblock Do large language models latently perform multi-hop reasoning?
\newblock \emph{arXiv preprint arXiv:2402.16837}.

\bibitem[{Yao et~al.(2024)Yao, Yu, Zhao, Shafran, Griffiths, Cao, and Narasimhan}]{yao2024tree}
Shunyu Yao, Dian Yu, Jeffrey Zhao, Izhak Shafran, Tom Griffiths, Yuan Cao, and Karthik Narasimhan. 2024.
\newblock Tree of thoughts: Deliberate problem solving with large language models.
\newblock \emph{Advances in Neural Information Processing Systems}, 36.

\bibitem[{Zhang et~al.(2021)Zhang, Raghu, Kleinberg, and Bengio}]{zhang2021pointer}
Chiyuan Zhang, Maithra Raghu, Jon Kleinberg, and Samy Bengio. 2021.
\newblock Pointer value retrieval: A new benchmark for understanding the limits of neural network generalization.
\newblock \emph{arXiv preprint arXiv:2107.12580}.

\bibitem[{Zhang et~al.(2023)Zhang, Zhang, Zhang, Liu, and Huang}]{zhang2023beam}
Jiahao Zhang, Haiyang Zhang, Dongmei Zhang, Yong Liu, and Shen Huang. 2023.
\newblock Beam retrieval: General end-to-end retrieval for multi-hop question answering.
\newblock \emph{arXiv preprint arXiv:2308.08973}.

\bibitem[{Zhang et~al.(2024{\natexlab{a}})Zhang, Yuan, and Yao}]{zhang2024diagram}
Yifan Zhang, Yang Yuan, and Andrew Chi-Chih Yao. 2024{\natexlab{a}}.
\newblock On the diagram of thought.
\newblock \emph{arXiv preprint arXiv:2409.10038}.

\bibitem[{Zhang et~al.(2024{\natexlab{b}})Zhang, Lin, Wang, Zhang, and Xu}]{zhang2024initialization}
Zhongwang Zhang, Pengxiao Lin, Zhiwei Wang, Yaoyu Zhang, and Zhi-Qin~John Xu. 2024{\natexlab{b}}.
\newblock Initialization is critical to whether transformers fit composite functions by inference or memorizing.
\newblock \emph{arXiv preprint arXiv:2405.05409}.

\bibitem[{Zhang et~al.(2024{\natexlab{c}})Zhang, Wang, Yao, Zhou, Li, Xu et~al.}]{zhang2024anchor}
Zhongwang Zhang, Zhiwei Wang, Junjie Yao, Zhangchen Zhou, Xiaolong Li, Zhi-Qin~John Xu, and 1 others. 2024{\natexlab{c}}.
\newblock Anchor function: a type of benchmark functions for studying language models.
\newblock \emph{arXiv preprint arXiv:2401.08309}.

\bibitem[{Zhou et~al.(2022)Zhou, Zhou, Jin, Luo, Zhang, and Xu}]{zhou2022empirical}
Hanxu Zhou, Qixuan Zhou, Zhenyuan Jin, Tao Luo, Yaoyu Zhang, and Zhi-Qin~John Xu. 2022.
\newblock Empirical phase diagram for three-layer neural networks with infinite width.
\newblock \emph{Advances in Neural Information Processing Systems}.

\bibitem[{Zhu et~al.(2018)Zhu, Wu, Yu, Wu, and Ma}]{zhu2018anisotropic}
Zhanxing Zhu, Jingfeng Wu, Bing Yu, Lei Wu, and Jinwen Ma. 2018.
\newblock The anisotropic noise in stochastic gradient descent: Its behavior of escaping from sharp minima and regularization effects.
\newblock \emph{arXiv preprint arXiv:1803.00195}.

\end{thebibliography}

\appendix

\newpage
\appendix
\onecolumn

\section{Related Work}

    \textbf{Language model reasoning} There have been numerous experimental and theoretical studies on language model reasoning. \cite{abbe2022learning} examines the reasoning capabilities of neural networks using the Pointer Value Retrieval (PVR) benchmark, which was originally introduced in \cite{zhang2021pointer}. \cite{sharma2023truth} demonstrates that applying low-rank approximation of certain weight layers to themselves can enhance reasoning performance across various tasks, and \cite{chen2024truncating} explains this phenomenon in a two-layer Transformer model. \cite{wang2024grokked} investigates both ID and OOD reasoning abilities on two synthetic tasks involving composition and comparison. \cite{jiang2024peek} reveals that the reasoning process in Large Language Models (LLMs) is influenced by token bias and that these models continue to face challenges when dealing with more complex logical tasks. \cite{abbe2024far} introduces the distribution locality and shows that Transformers require low locality. \cite{zhang2024initialization, luo2021phase, zhou2022empirical} shows that small initialization can facilitate model reasoning. \citet{brinkmann2024mechanistic} investigates the mechanism by which language models output intermediate reasoning paths in multi-step reasoning tasks. \blue{\citet{aubry2024transformer} uncover a transformer block coupling phenomenon in a variety of LLMs by tracing the trajectories of individual tokens as they pass through transformer blocks.} \red{Recently, the multi-hop reasoning abilities of models have garnered attention in the LLM field. \citet{kil2024ii, li2024making} promote the use of Chain-of-Thought (CoT) reasoning by models to perform multi-step thinking through appropriate prompt engineering. \citet{zhang2023beam} proposes a new retrieval framework for multiple reasoning paths. \citet{dar2022analyzing, li2024understanding, yang2024large, shalev2024distributional} pinpoint where models execute multi-step reasoning by causal intervention and observing neuron activation. \citet{biran2024hopping} introduces a method to enhance models' reasoning abilities by repeatedly invoking intermediate layers. These works, especially the causal intervention experiments, have inspired our research. However, existing studies primarily conduct macro-level statistical analyses on actual complex large language models. While causal intervention methods can help us identify critical paths, our work builds upon this foundation by further exploring how models leverage their own weights to generate these key paths. Experiments on our symbolic datasets also facilitate more in-depth experimental and theoretical investigations.}
    
    \textbf{Understanding the mechanism of neural model} Our work builds upon previous studies on the attention mechanism \citep{voita2019analyzing,vig2019multiscale,kovaleva2019revealing,kobayashi2020attention}. Numerous researchers have proposed various approaches to identify the roles of different heads in Transformers \citep{vig2020investigating,jeoung2022changed,wang2022interpretability,conmy2023towards,merullo2023circuit,guo2023transformers,wang2024understanding,amsel2024benefits,li2024chain, wang2024understanding1}. These methods predominantly employ the concept of perturbation. Similar to the observations made by \citet{wang2023label} and \citet{dutta2024think}, who noted that large language models typically perform information aggregation in shallow layers and information induction in deeper layers, we have also observed comparable phenomena in our study. The idea of symbolic datasets is inspired by \citet{poli2024mechanistic,zhang2024anchor}. There have also been some insightful theoretical works on feedforward neural networks. A series of studies have explored neural network preferences and generalization from the perspectives of regularization and frequency, etc. \citep{xu2019frequency, wang2024improving, jacot2018neural, jacot2020implicit, arora2019fine, arora2018convergence}.  And \citet{wu2023implicit, wang2310theoretical, arora2022understanding, li2021validity, wu2018sgd, zhu2018anisotropic, arora2019exact} investigates the dynamical behavior of neural networks, while \citet{ren2024understanding} examines the factors influencing neural network generalization.

    \textbf{In-context learning and induction head} Our work primarily investigates the model's ability to perform in-context learning. The concept of in-context learning (ICL) was first introduced by \citet{brown2020language}. Since then, a series of studies have utilized induction heads to investigate ICL, yielding remarkable research outcomes \citep{olsson2022context, garg2022can, wang2022interpretability, muller2021transformers, goldowsky2023localizing, bietti2024birth, nichani2024transformers, edelman2024evolution, chen2024training, todd2023function, chen2024can}. It is worth noting that induction heads can be considered as a special case of multi-step reasoning tasks with the reasoning step equal to $1$. However, multi-step reasoning is not a simple linear combination of single-step reasoning. In our work, we study the mechanism that enables multi-step reasoning, which has not been explored in previous studies.

\section{Experimental settings}\label{appendix: settings}

    In our experiments (Section~\ref{VTS}), the vocabulary size is set to $d=201$, and the hidden space dimension is set to $d_m=400$ and $d_q=d_k=d_v=64$. We use 200,000 2-step reasoning sequences (with sequence length equal to 13). The learning rate is set to 2e-5 and linearly warms up to 1e-4 within 400 epochs and then decays to 1e-5 within 3600 epochs. The batch size is set to 100. We use the AdamW optimizer with default parameters as set in PyTorch 2.3.0. 

    The experiments were conducted on a server with the following configuration:
    \begin{itemize}[topsep=0pt, partopsep=0pt, itemsep=0pt, parsep=0pt]
        \item 64 AMD EPYC 7742 64-Core Processor.
        \item 256GB of total system memory.
        \item 2 NVIDIA A100 GPUs with 40GB of video memory each and 8 NVIDIA GeForce RTX 4080 GPUs with 16GB of video memory each.
        \item The experiments were run using the Ubuntu 22.04 LTS operating system.
    \end{itemize}
    The task shown above can be completed within 2 hours with a single NVIDIA A100 GPU. For other, more complex examples, they can be finished within 24 hours.

\section{Proof of Lemma~\ref{lemma1}} \label{proof}

    \textbf {Lemma 1} \textit{
        Suppose token $\vx=\sum_{i=1}^n \va_i \mW_i \in \mathbb{R}^{d_m}$ and token $\vy=\sum_{i=1}^n \vb_i \mW_i \in \mathbb{R}^{d_m}$, where $\va_i,\vb_i\in \mathbb{R}^{d_m}$, $\mW_i\in \mathbb{R}^{d_m\times d_m}, \ i=1,2,\cdots, n$. Each element of $\{\va_i\}_{i=1}^n$, $\{\vb_i\}_{i=1}^n$ and $\{\mW_i\}_{i=1}^n$ follows a normal distribution $\mathcal{N}(0,1/d_m)$ and independent to others. Then, we have: 
        \begin{equation}
            \vx \mW_i^{\mathsf{T}} = \va_i + \mathcal{O}\left(\sqrt{\frac{n}{d_m}}\right),\quad
            \vy \mW_j^{\mathsf{T}} = \vb_j + \mathcal{O}\left(\sqrt{\frac{n}{d_m}}\right),
        \end{equation}
        \begin{equation}
            \vx \mW_i^{\mathsf{T}} \mW_j \vy^{\mathsf{T}} = \va_i \vb_j^{\mathsf{T}} + \mathcal{O}\left(\frac{n}{\sqrt{d_m}}\right).
        \end{equation}
    }

    \begin{proof}
        We first show that $\mW_i\mW_j^\mathsf{T}$ (denoted as $\mZ^{(i,j)}$) is also a random matrix with elements following a normal distribution $\mathcal{N}(0,1/d_m)$ when $i\ne j$. In fact,
        \begin{equation*}
            \E[(\mW_i\mW_j^\mathsf{T})_{s,t}] = \E[\sum_{k=1}^{d_m}(\mW_i)_{sk}(\mW_j)_{kt}]=\sum_{k=1}^{d_m}\E[(\mW_i)_{sk}]\E[(\mW_j)_{kt}]=0,
        \end{equation*}
        \begin{align*}
            &\Var\left[(\mW_i\mW_j^\mathsf{T})_{s,t}\right] = \E\left[\left(\sum_{k=1}^{d_m}(\mW_i)_{sk}(\mW_j)_{kt}\right)^2\right]\\
            &=\sum_{k=1}^{d_m}\E\left[(\mW_i)_{sk}^2\right]\E\left[(\mW_j)_{kt}^2\right]
            =d_m\times(\frac{1}{d_m})^2=\frac{1}{d_m},
        \end{align*}
        which indicate $\{\mZ^{(i,j)}\}$ follows the same distribution as $\{\mW_i\}_{i=1}^n$. Therefore, 

        \begin{align*}
            \E_{\{\mW_j\}}\left[(\vx\mW_i^\mathsf{T})_{t}\right] &= \sum_{j=1}^n\E_{\{\mZ^{(j,i)}\}_j}\left[(\va_j\mZ^{(j,i)})_t\right] \\
            &= \sum_{k=1}^{d_m}(\va_i)_{k}\E\left[(\mZ^{(i,i)})_{kt}\right] + \sum_{\substack{j=1\\j\ne i}}^n\sum_{k=1}^{d_m}(\va_j)_{k}\E\left[(\mZ^{(j,i)})_{kt}\right]\\
            &=\sum_{k=1}^{d_m}(\va_i)_{k}\E\left[(\mZ^{(i,i)})_{kt}\right]=(\va_i)_{t},
        \end{align*}

        \begin{align*}
            \Var_{\{\mW_j\}}\left[(\vx\mW_i^\mathsf{T})_{t}\right] &= \sum_{j=1}^n\Var_{\{\mZ^{(j,i)}\}_j}\left[(\va_j\mZ^{(j,i)})_t\right] \\
            &= \sum_{k=1}^{d_m}(\va_i)_{k}^2\Var\left[(\mZ^{(i,i)})_{kt}\right] + \sum_{\substack{j=1\\j\ne i}}^n\sum_{k=1}^{d_m}(\va_j)_{k}^2\Var\left[(\mZ^{(j,i)})_{kt}\right]\\
            &=\sum_{k=1}^{d_m}(\va_i)_{k}^2\Var\left[(\mZ^{(i,i)})_{kt}\right]+
            \frac{1}{d_m}\sum_{\substack{j=1\\j\ne i}}^n\sum_{k=1}^{d_m}(\va_j)_{k}^2\\
            &=(\va_i)_{t}^2\Var\left[(\mZ^{(i,i)})_{tt}\right]+\sum_{\substack{k=1\\k\ne t}}^{d_m}(\va_i)_{k}^2\Var\left[(\mZ^{(i,i)})_{kt}\right]+
            \frac{1}{d_m}\sum_{\substack{j=1\\j\ne i}}^n\sum_{k=1}^{d_m}(\va_j)_{k}^2\\
            &=\frac{2}{d_m}(\va_i)_{t}^2+\frac{1}{d_m}\sum_{\substack{k=1\\k\ne t}}^{d_m}(\va_i)_{k}^2+
            \frac{1}{d_m}\sum_{\substack{j=1\\j\ne i}}^n\sum_{k=1}^{d_m}(\va_j)_{k}^2\\
            &=\frac{1}{d_m}(\va_i)_{t}^2+ \frac{1}{d_m}\sum_{j=1}^n\sum_{k=1}^{d_m}(\va_j)_k^2.
        \end{align*}
        Therefore, $\Var\left[(\vx\mW_i^\mathsf{T})_{t}\right]=\Var_{\va}\left[\Var_{\{\mW_j\}}\left[(\vx\mW_i^\mathsf{T})_{t}\right]\right]=n/d_m+1/d_m^2$. And Chebyshev's inequality implies that $\vx \mW_i^{\mathsf{T}} = \va_i + \mathcal{O}\left(\sqrt{\frac{n}{d_m}}\right)$, which also holds for $\vy \mW_j^{\mathsf{T}}$.

        Assume that $\vx \mW_i^{\mathsf{T}} = \va_i+\vz_1,\ \vy \mW_j^{\mathsf{T}}=\vb_j+\vz_2$, $\vz_1,\vz_2$ are random vector with the elements follow the normal distribution $\mathcal{N}(n,1/d_m)$, then,
        \begin{align*}
            &\Var\left[\vx \mW_i^{\mathsf{T}} \mW_j \vy^{\mathsf{T}}\right] 
            = \Var\left[\va_i \vb_j^{\mathsf{T}} +\va_i\vz_2^\mathsf{T}+\vz_1\vb_j^{\mathsf{T}}+\vz_1\vz_2^{\mathsf{T}}\right]\\
            &=d_m\cdot\frac{1}{d_m}\cdot\frac{1}{d_m}+2\cdot d_m\cdot \frac{1}{d_m}\cdot\frac{n}{d_m}+d_m\cdot\frac{n}{d_m}\cdot\frac{n}{d_m} =\frac{(n+1)}{d_m}.
        \end{align*}

        Thus we have $\vx \mW_i^{\mathsf{T}} \mW_j \vy^{\mathsf{T}} = \va_i \vb_j^{\mathsf{T}} + \mathcal{O}\left(\frac{n}{\sqrt{d_m}}\right)$.
    \end{proof}

\clearpage

\section{Further Discussion on the Vertical Thinking Strategy}\label{appendix: Matching Operation}

    \subsection{Information Fusion}
    
\begin{figure}[htb]
        \centering
        \includegraphics[width=0.72\textwidth]{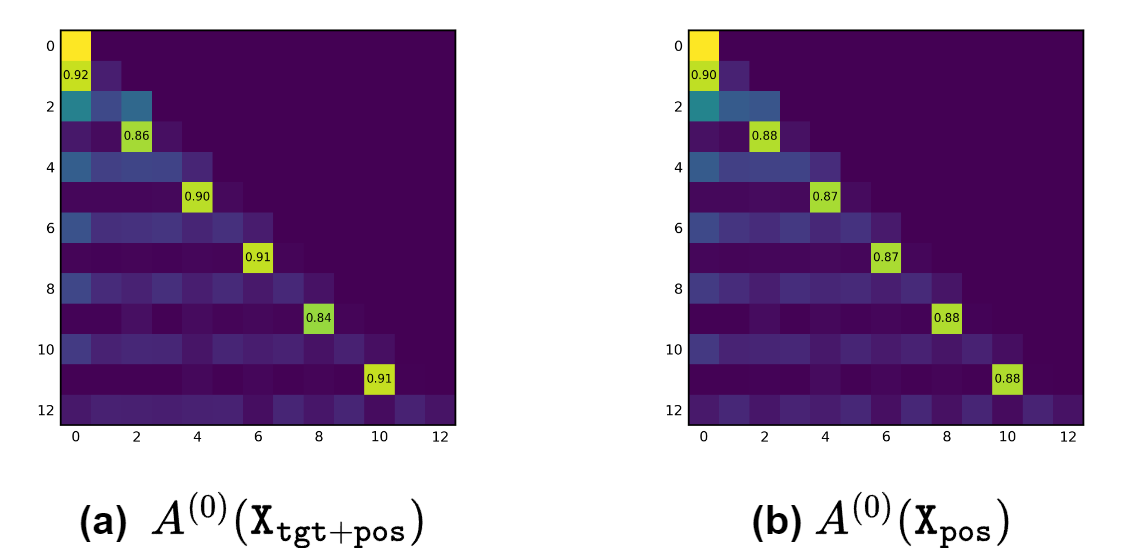}
        \caption{A comparison between the original attention (a) and the positional attention (b) of the first Transformer block, where $\mX_{tgt+pos}=\mX_{tgt}+\mX_{pos}$.}
        \label{fig:pos}
    \end{figure}

    In our symbolic dataset task, as mentioned in Section~\ref {VTS}, the first layer facilitates the fusion of token information at odd and even positions. We find that positional encoding plays a crucial role in the features of the first layer of attention. Fig.~\ref{fig:pos}(a)(b) illustrates a comparison between the original attention mechanism and the positional attention mechanism calculated with eq.~\ref{eq:pos attn}. As shown, there is minimal difference between the two approaches.
    
    \begin{equation}
        \fA^{(0)}(\mX^{\text{pos}}) = \mathrm{softmax}\left(\frac{\mathrm{mask}(\mX^{\text{pos}}\mW^{qk}\mX^{\text{pos},\mathsf{T}})}{\sqrt{d_k}}\right).
        \label{eq:pos attn}
    \end{equation}

    \subsection{Detailed Matching Matrix}

    In Section~\ref {VTS}, for simplicity of analysis, we ignored the impact of the feedforward layer. Here, we define a detailed version of the matching matrix as follows:
    \begin{align}
        \tilde{h}^{(1)}(\mX)&=f^{(0)}(\mX)\mW^{q(1),\mathsf{T}}[f^{(0)}(\mX\mW^{vo(0)})\mW^{k(1),\mathsf{T}}]^\mathsf{T}\\
        \tilde{h}^{(2)}(\mX) &= f^{(1)}[f^{(0)}(\mX)\mW^{vo(1)}]\mW^{q(2),\mathsf{T}} \left[f^{(1)}\circ f^{(0)}(\mX\mW^{vo(0)})\mW^{k(2),\mathsf{T}}\right]^\mathsf{T}.
    \end{align}

\begin{figure}[htb]
        \centering
        \includegraphics[width=0.82\textwidth]{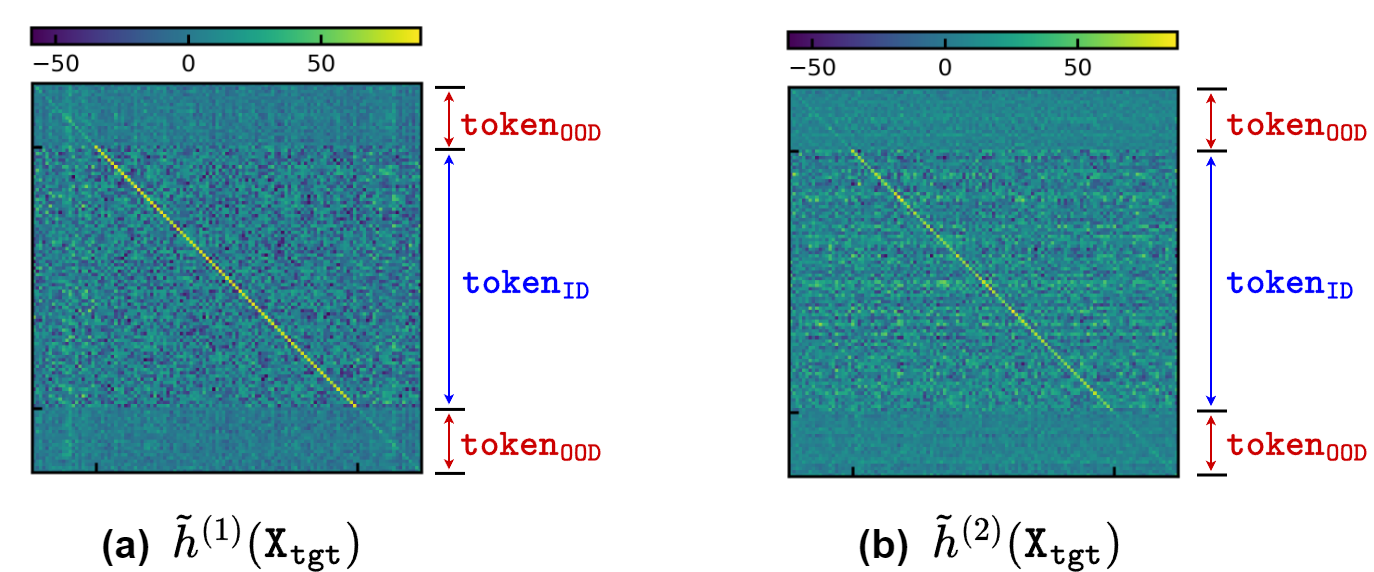}
        \caption{(a) Heatmap of $\tilde{h}^{(1)}(\mX_{tgt})$ and $\tilde{h}^{(2)}(\mX_{tgt})$. The diagonal elements exhibit the largest values, confirming the matching operation.}
        \label{fig:appendix_matching}
    \end{figure}

    As shown in Fig.~\ref{fig:appendix_matching}, the detailed matching matrices still maintain the diagonal element property in most cases, even for the out-of-distribution tokens. 

    \subsection{\blue{Independence of Buffers}}

\begin{figure}[htb]
        \centering
        \includegraphics[width=0.4\textwidth]{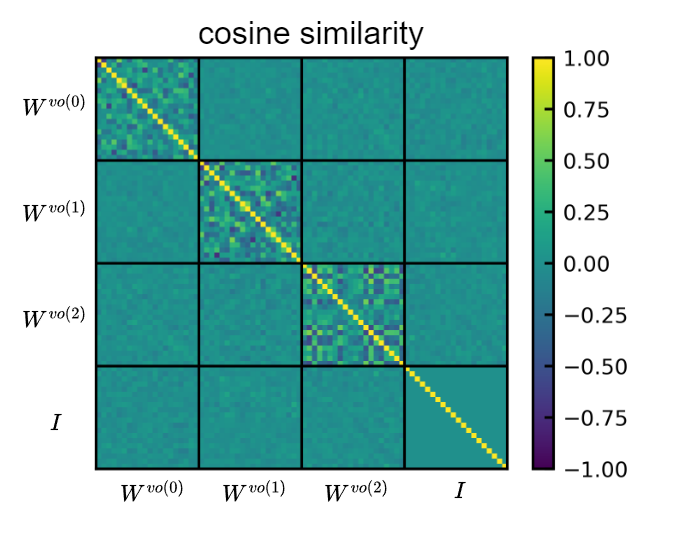}
        \caption{\blue{The cosine similarity between row vectors of different buffers ($\mW^{vo(0)},\mW^{vo(1)},\mW^{vo(2)}$ and $\mI$).}}
        \label{fig:buffer_cos}
    \end{figure}

    \blue{To verify the independence between buffers, we computed and visualized the cosine similarity between row vectors of different buffers ($\mW^{vo(0)},\mW^{vo(1)},\mW^{vo(2)}$ and $\mI$). As shown in Fig.~\ref{fig:buffer_cos}, apart from exhibiting a certain similarity within itself (so-called condense phenomenon\citep{luo2021phase}), each buffer remains nearly orthogonal to the others.}

    \subsection{Weight Construction Method for Multi-Step Reasoning Networks}

    In Section~\ref {VTS}, we mentioned that by setting the weights in the following manner, we can enable an $L$-layer Transformer model to possess $(L-1)$-step reasoning capability.
    \begin{equation}
        \mW^{q(0)}=\mW^{q(1)}=I,\  \mW^{q(l)}=\mW^{vo(l-1),\mathsf{T}},\ l\ge 2, 
    \end{equation}
    \begin{equation}
        \mW^{k(0)}=\sum_{i=1}^{[l_\text{seq}/2]}\vp_{2i}\vp_{2i-1}^{\mathsf{T}}, \ \mW^{k(l)}=\mW^{vo(0),\mathsf{T}},\ l\ge 1,
    \end{equation}
    where $\{\mW^{vo(l)}\}_{l=1}^L$ are random matrices and the projection weight $\mW^{p} = \mW^{vo(L),\mathsf{T}} \mW^{\text{emb},\mathsf{T}}$.
    
\begin{figure}[htb]
        \centering
        \includegraphics[width=\textwidth]{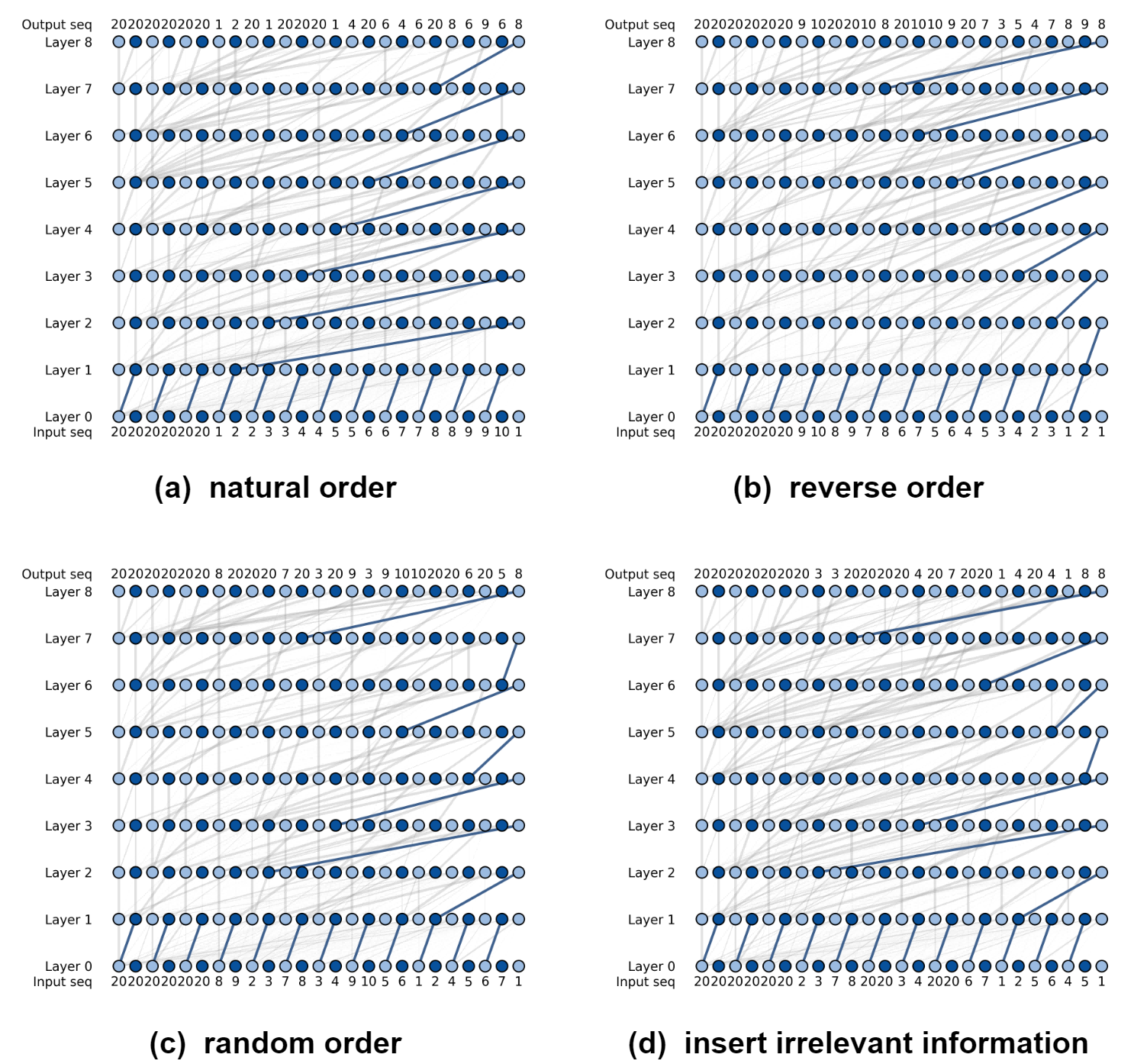}
        \caption{The test results for the 8-layer Transformer we constructed. The gray lines represent attention values that do not affect the final outcome. The width of all lines is positively correlated with the attention weights.}
        \label{fig:induction_ini}
    \end{figure}

    The specific construction method is as follows: To ensure that each buffer in the model has adequate robustness against interference, we set $d_m = d_q = d_k = d_v = 10000$. The feedforward layers are assigned zero weights so that the residual connection dominates. Since all the weight matrices we use are untrained random matrices, the layer normalization will have no effect. Fig.~\ref{fig:induction_ini} shows the multi-step reasoning ability of an 8-layer Transformer. We tested natural order, reverse order, random order sentences, and sentences with inserted irrelevant tokens (i.e., token \texttt{[20]}), and the model was able to output the correct answer \texttt{[8]} in all cases. Each sentence begins with token \texttt{[20]} to prevent $\fA^{(l)}_{0,0}$ from always equaling 1, which could affect the buffer.
    
\clearpage

\section{Details for RMBA Experiment} \label{appendix: add noise}

    This section provides supplementary details on the experimental setup for the RMBA experiment. We use a 3-layer single-head Transformer with $d_q=d_k=d_v=64$ and $d_m=400$. The training set consists of 30,000 2-step reasoning chains. We trained the model for 200 epochs in total. In this setting, the Transformers have poor generation ability even in the in-distribution test dataset. 
    
\begin{figure}[htb]
        \centering
        \includegraphics[width=0.8\textwidth]{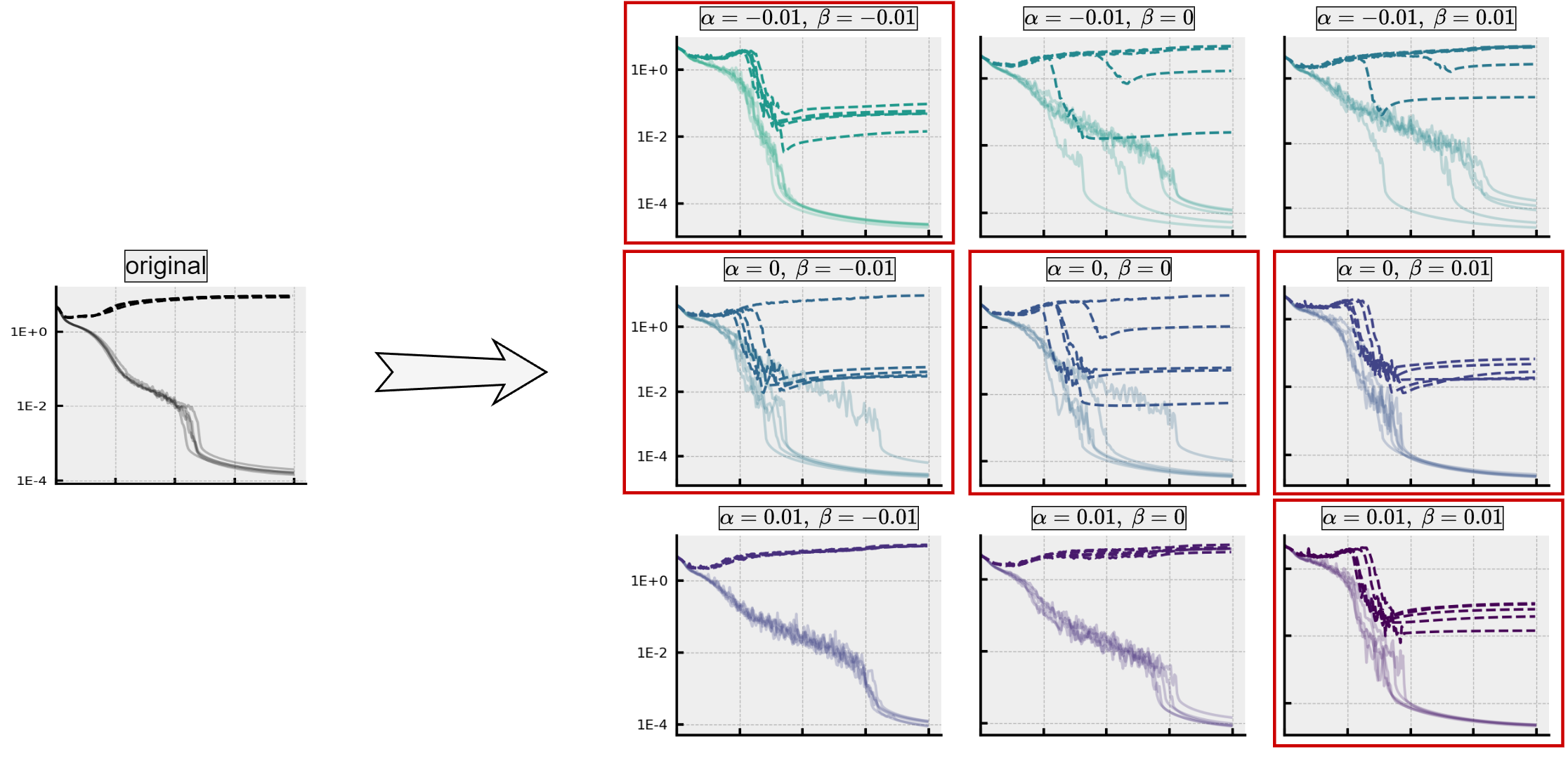}
        \caption{The impact of different learnable parameters' initial values, $\alpha^{(l)}$ and $\beta^{(l)}$, on the model's reasoning ability. The solid lines represent the training loss, while the dashed lines denote the test loss. Each experiment was conducted with 5 random seeds.}
        \label{fig:add noise loss}
    \end{figure}

\begin{figure}[htb]
        \centering
        \includegraphics[width=\textwidth]{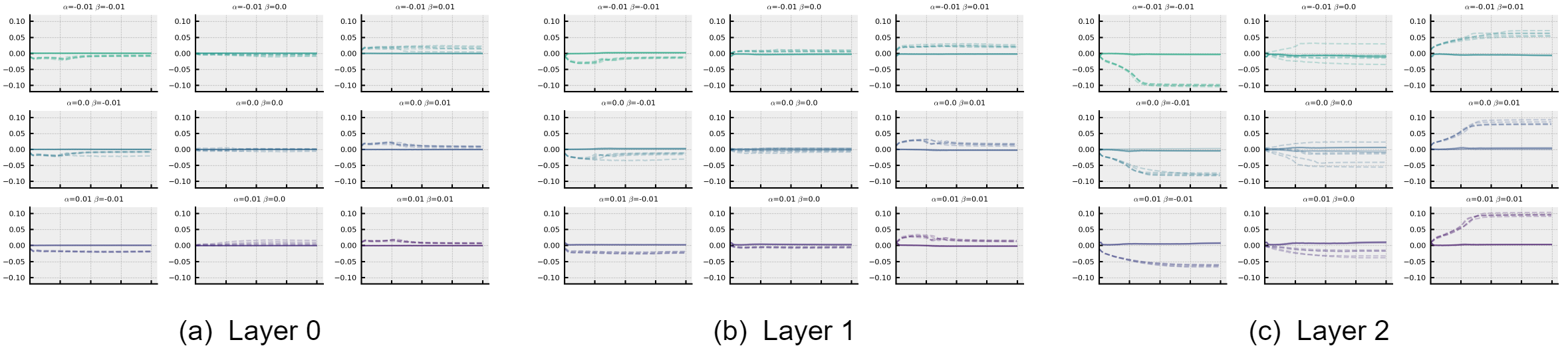}
        \caption{Changes of the learnable parameters $\alpha^{(l)}$ and $\beta^{(l)}$ during training. The solid lines represent the $\alpha^{(l)}$, while the dashed lines denote the $\beta^{(l)}$. Each experiment was conducted with 5 random seeds.}
        \label{fig:add noise ab}
    \end{figure}

    We replace $\mW^{qk(l)}$ and $\mW^{vo(l)}$ with $\mW^{qk(l)}+\alpha^{(l)}\mZ^{(l-1)}$ and $\mW^{vo(l)}+\beta^{(l)}\mZ^{(l)}$, respectively, where $\alpha^{(l)}$ and $\beta^{(l)}$ are learnable parameters, and $\{\mZ^{(l)}\}_{l=1}^L$ is a set of random matrix following $N(0,1/d_m)$. Therefore, 6 extra learnable parameters are added to this 3-layer single-head model in total. Fig.~\ref{fig:add noise loss} shows the loss of the Transformer under different settings. Fig.~\ref{fig:add noise ab} shows the changes of the learnable parameters $\alpha^{(l)}$ and $\beta^{(l)}$ during training.

\begin{figure}[htb]
        \centering
        \includegraphics[width=0.8\textwidth]{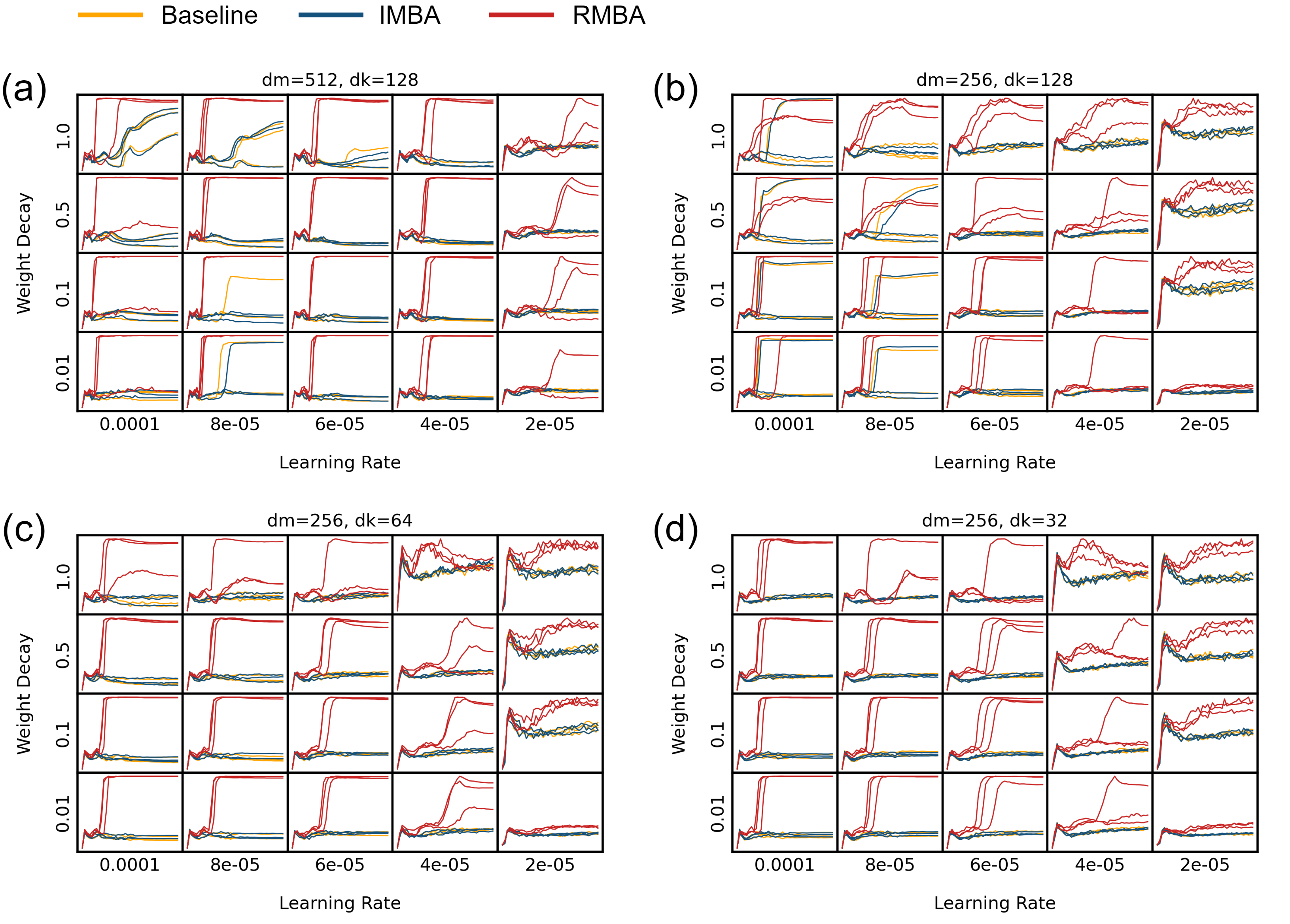}
        \caption{\red{A comparison of the training results for RMBA, IMBA, and the Baseline model under different hyperparameters is presented. We investigated learning rates ranging from 2e-5 to 1e-4 and weight decay values from 0.01 to 1. We considered four configurations of the hidden space dimension. For each hyperparameter setting, we conducted experiments using 3 different random seeds, totaling 720 experiments.}}
        \label{fig:compare specific}
    \end{figure}

    We further tested the performance of the three algorithms under different hyperparameter configurations. We investigated the effects of weight decay, learning rate, and hidden dimension on training. For the RMBA and IMBA algorithms, we set $\alpha^{(l)}_{\text{ini}}$=$\beta^{(l)}_{\text{ini}}$=0.01. As shown in Fig.~\ref{fig:compare specific}, under various settings, the RMBA algorithm consistently facilitated the model's ability to generalize.

    \begin{figure}[htb]
        \centering
        \includegraphics[width=\textwidth]{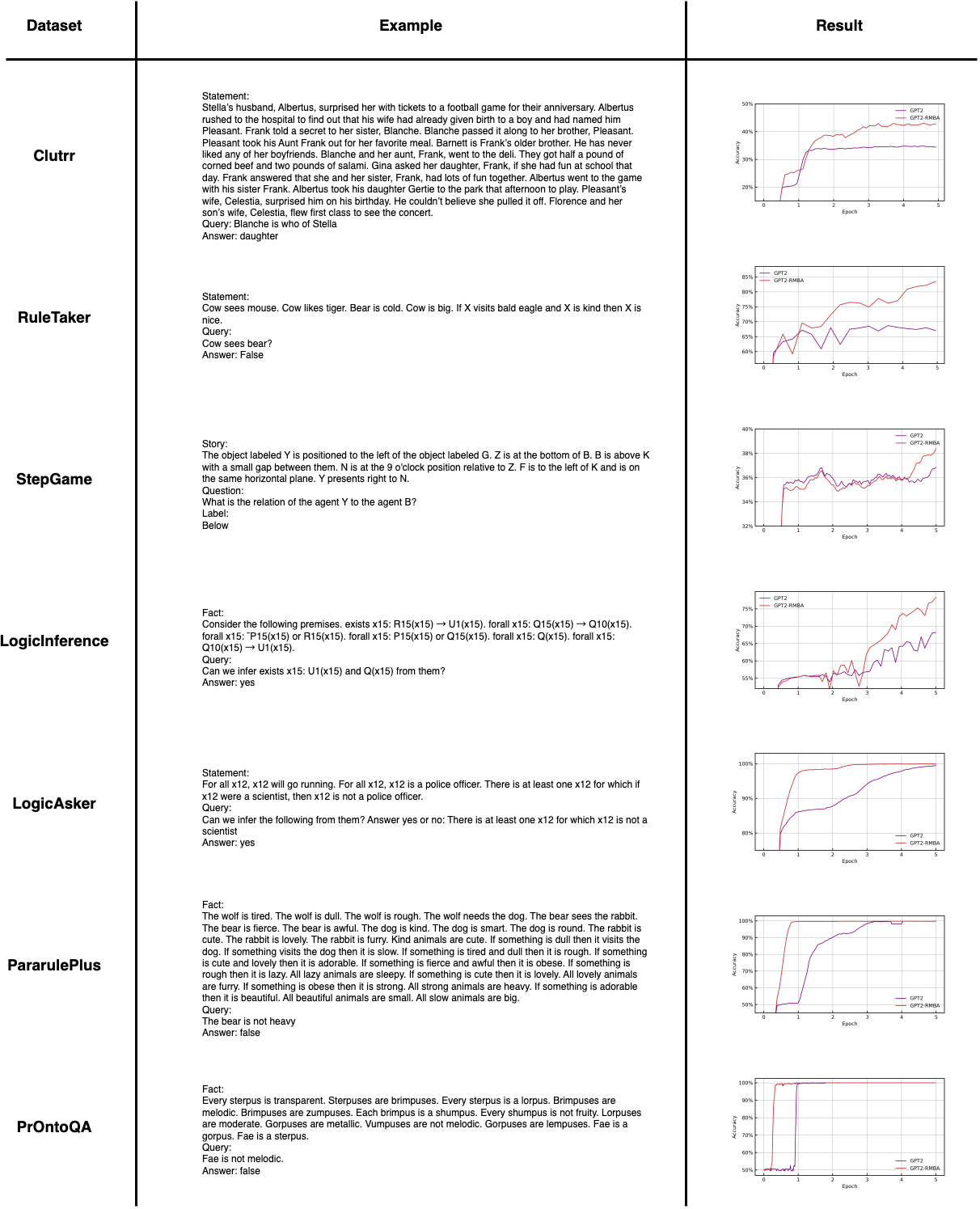}
        \caption{Examples from each dataset, along with their training curves.}
        \label{fig:all_dataset_acc}
    \end{figure}

\clearpage

\section{Details for Horizontal Thinking Strategy}\label{appendix: CoT}

We trained a 2-layer Transformer model and investigated its ability to perform multi-step reasoning with horizontal thinking. Specifically, we trained the Transformer model on 20,000 samples of length 13, each labeled with the result of a one-step reasoning process. 

\begin{figure}[htb]
        \centering
        \includegraphics[width=0.8\textwidth]{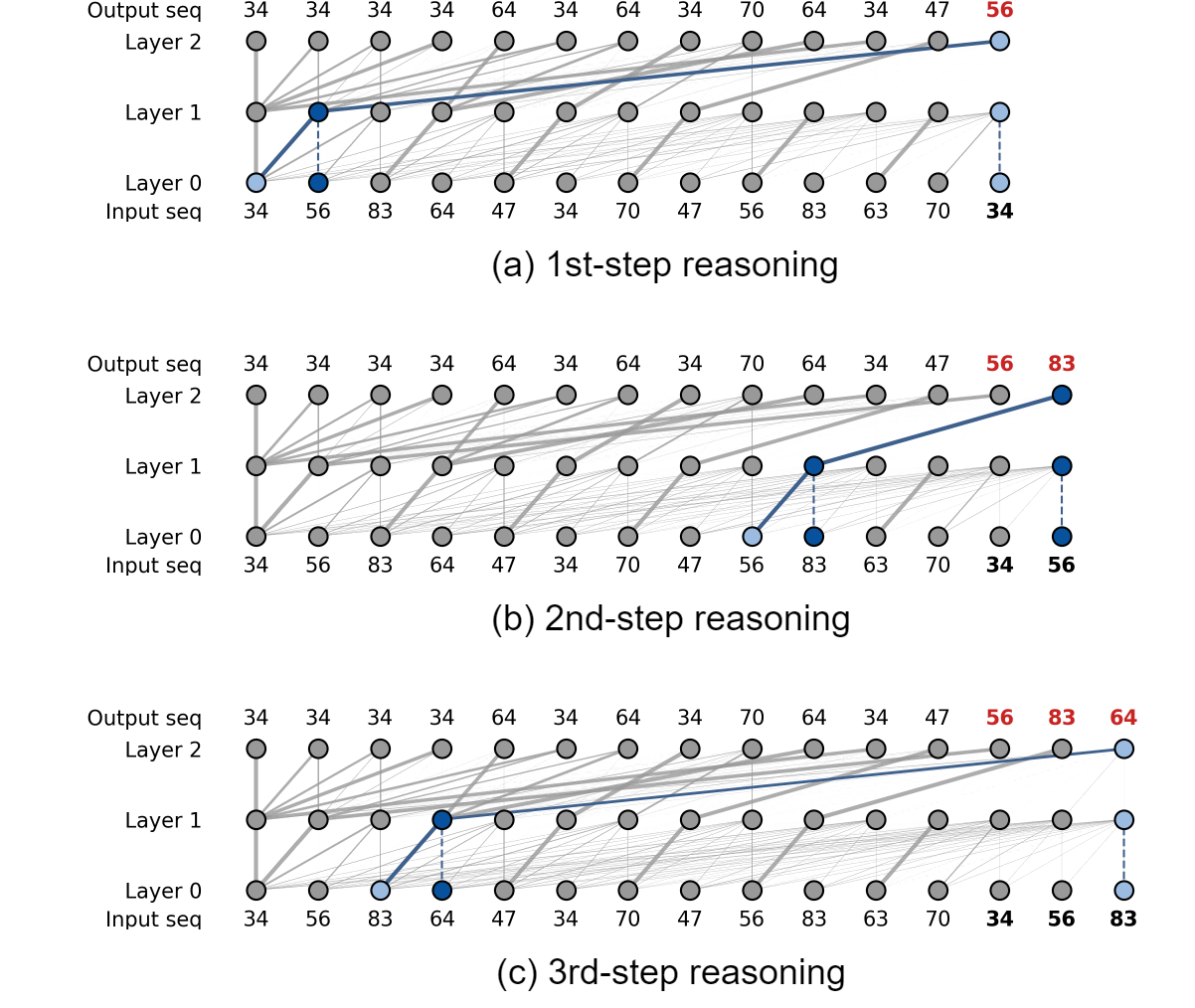}
        \caption{\red{An illustration of the process of performing 3-step reasoning using a 2-layer Transformer model with CoT. The width of the connections in the diagram is based on the attention weights.}}
        \label{fig:CoT example}
    \end{figure}

As shown in Fig.~\ref{fig:CoT example}, when we fed the model's output back into the model, it was able to generate the next step's reasoning result, even though it had never been exposed to sentences longer than length 13 during training. 

\clearpage

\section{Matching Score and Kernel Score for Real World LLMs}\label{appendix: Phi-3}

    In this section, we calculate the matching score and kernel score of the large language model Phi-3\citep{abdin2024phi}. 
    
\begin{figure}[htb]
        \centering
        \includegraphics[width=0.9\textwidth]{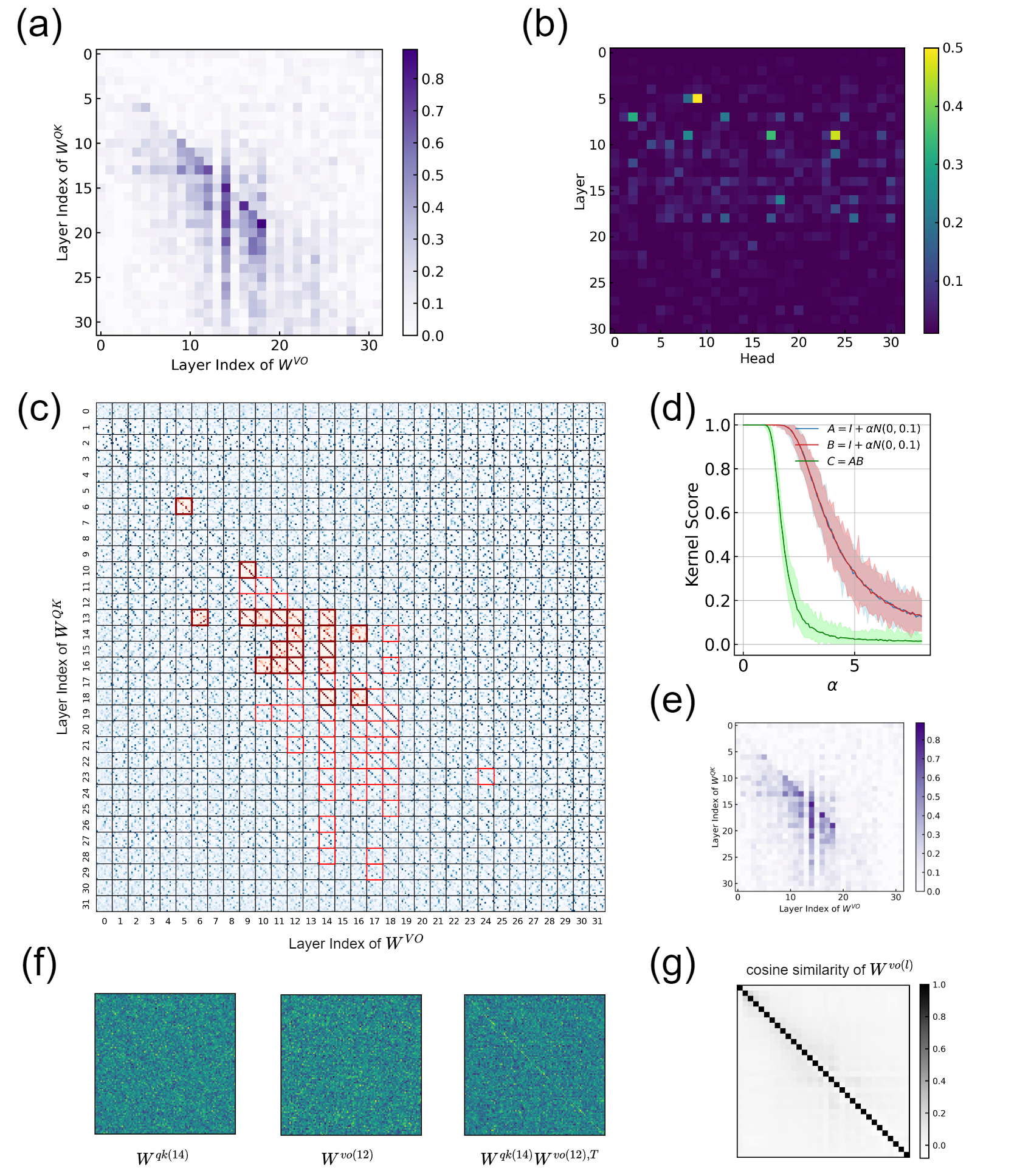}
        \caption{\red{The calculation results of the kernel score and matching score for the Phi-3 model. (a) Visualization of $\sigma(\text{Ker}^{(l_1, l_2)})$. (b) Visualization of the matching score calculations for each head in each layer, indicating that reasoning is concentrated in the middle layers of the model. (c) Visualization of the kernel matrix between layers, where the subgraphs enclosed in red boxes correspond to $\text{KS}(\text{Ker}) > 0.3$ and the subgraphs enclosed in darkred boxes correspond to $\text{KS}(\text{Ker}) > 0.3$ but $\max(\text{KS}(\sum_h \mW^{q(h)}\mW^{k(h),\mathsf{T}}), \text{KS}(\mW^{vo,\mathsf{T}})) < 0.3$. }\blue{(d) The kernel score of a noise-added identity matrix(A and B) and the kernel score of the product of two noise-added identity matrices(C). It can be observed that for two unrelated matrices, when their individual kernel scores are less than 0.3, the kernel score of their product is less than 0.025. (e) The kernel score obtained after setting the diagonal elements of $\sum_h \mW^{q(h)}\mW^{k(h),\mathsf{T}}$ and $\mW^{vo,\mathsf{T}}$ to zero. (f) Visualization of the $\sum_h \mW^{q(14, h)}\mW^{k(14, h),\mathsf{T}}$ and the $\mW^{vo(l_2),\mathsf{T}}$ in the Phi-3 model, along with their inner product. Despite their weak diagonal structure individually, their inner product exhibits a clear diagonal structure.} \green{(g) Cosine similarity of flattened $\mW^{vo(l_1)}$ and $\mW^{vo(l_2)}$, $l_1,l_2 \in \{0,\cdots, 31\}$.}}
        \label{fig:phi-3}
    \end{figure}

    We focus on whether $\mW^{vo}$ and $\mW^{qk}$ function as the information buffer and the information extractor, respectively. To simplify our analysis, we temporarily disregard the effects of the feedforward layers. Following the method described in the main text, we compute $\text{Ker}^{(l_1, l_2)} = \mW^{qk(l_1)} \mW^{vo(l_2),\mathsf{T}}$ and observe whether it exhibits a dominant diagonal characteristic. For multi-head models, the above equation is modified as:
    \begin{equation}
        \text{Ker}^{(l_1, l_2)} = \left( \sum_h \mW^{q(l_1, h)}\mW^{k(l_1, h),\mathsf{T}} \right) \mW^{vo(l_2),\mathsf{T}}.
    \end{equation}
    We define the Kernel Score (KS) as 
    \begin{equation}
        \text{KS}(\text{Ker}^{(l_1, l_2)}) = \text{Trace}(\sigma(\text{Ker}^{(l_1, l_2)}))/d_m,
    \end{equation}
    which measures the ability of layer $l_1$ in the model to extract information cached at layer $l_2$. As shown in Fig.~\ref{fig:phi-3}(a), when $l_1 \geq l_2$, the kernel score is nearly zero, which aligns with the logical sequence of information processing. When $l_1 < l_2$, the kernel score decreases as $(l_2-l_1)$ increases, indicating that the model tends to extract the most recently acquired information for further processing. In Fig.~\ref{fig:phi-3}(b), we plot $\sigma(\text{Ker}^{(l_1, l_2)})$ and highlight the regions where the Kernel Score$>0.3$.

    \blue{To further verify that the diagonal structure of $\text{Ker}^{(l_1, l_2)}$ arises from the alignment of model weights rather than the intrinsic diagonal structure of $\sum_h \mW^{q(l_1, h)}\mW^{k(l_1, h),\mathsf{T}}$ and $\mW^{vo(l_2),\mathsf{T}}$, we conducted a small experiment, as illustrated in Fig.~\ref{fig:phi-3}(d). In this experiment, we assume that both matrices $A$ and $B$ are noise-added identity matrices, where the noise scale is denoted by $\alpha$. We then compute the kernel scores of $A$, $B$, and their product $C = AB$. The results show that when $\max(\text{KS}(A), \text{KS}(B)) < 0.3$, the Kernel score of the product, $\text{KS}(C)$, is less than 0.025. However, as shown in Fig.~\ref{fig:phi-3}(c), we observe that for many heads, even when $\max(\text{KS}(\sum_h \mW^{q(h)}\mW^{k(h),\mathsf{T}}), \text{KS}(\mW^{vo,\mathsf{T}})) < 0.3$, the $\text{Ker}^{(l_1, l_2)}$ still has a large kernel score. This indicates that the alignment of model weights is the key factor driving the diagonal structure of $\text{Ker}^{(l_1, l_2)}$. In Fig.~\ref{fig:phi-3}(f), we present an example where both $\sum_h \mW^{q(l_1, h)}\mW^{k(l_1, h),\mathsf{T}}$ and $\mW^{vo(l_2),\mathsf{T}}$ appear relatively disordered individually, but their product exhibits a clear diagonal structure.}

    \blue{Another straightforward method is to directly set the diagonal elements of $\sum_h \mW^{q(l_1, h)}\mW^{k(l_1, h),\mathsf{T}}$ and $\mW^{vo(l_2),\mathsf{T}}$ to zero, and then calculate the kernel score based on the resulting $\tilde{\text{Ker}}^{(l_1, l_2)}$. Fig.~\ref{fig:phi-3}(e) illustrates this result, showing that it is approximately the same as that in Fig.~\ref{fig:phi-3}(a).}

    Moreover, without loss of generality, we consider the case that includes LayerNorm(LN) and feedforward(FNN) layers. We compute the matching score for each head in each layer, with the specific calculation formula as follows:
    \begin{align}
    \mX^{vo} &= \text{LN}^{(l-1)}_{\text{attn}}(\mX) \mW^{v(l-1), \mathsf{T}} \mW^{o(l-1), \mathsf{T}},\label{multi-head vo}\\
    \mX^{vof} &= \mX^{vo} + \text{FNN}^{(l-1)}(\text{LN}^{(l-1)}_{\text{FNN}}(\mX^{vo})),\\
    \mX^{vok(h)} &= \text{LN}^{(l)}_{\text{attn}}(\mX^{vof}) \mW^{k(l, h)}, \\
    \mX^{f} &= \text{LN}^{(l-1)}_{\text{attn}}(\mX) + \text{FNN}^{(l-1)}(\text{LN}^{(l-1)}_{\text{FNN}}(\text{LN}^{(l-1)}_{\text{attn}}(\mX))),\\
    \mX^{q(h)} &= \mX^{f} \mW^{q(l, h)}, \\
    \text{matching matrix: } h^{(l,h)}(\mX) &= \mX^{q(h)} \mX^{vok(h), \mathsf{T}}.
    \end{align}
    We visualized the matching score of each head in each layer (Fig.~\ref{fig:phi-3}(b)) and found that the matching scores were highest in layers 5 to 20. This aligns with the conclusion mentioned in \cite{dutta2024think}, namely that the reasoning layers of large language models generally appear in the middle portion.

    \green{To further validate that Phi-3 might employ a buffer mechanism, we computed the pairwise cosine similarity of the matrices $\{\mW^{vo(l)}\}$ (each matrix is flattened as a long vector). The results indicate that these matrices are nearly orthogonal to each other, suggesting that they can be treated as independent buffers.}

    \clearpage

\begin{figure}[htb]
        \centering
        \includegraphics[width=\textwidth]{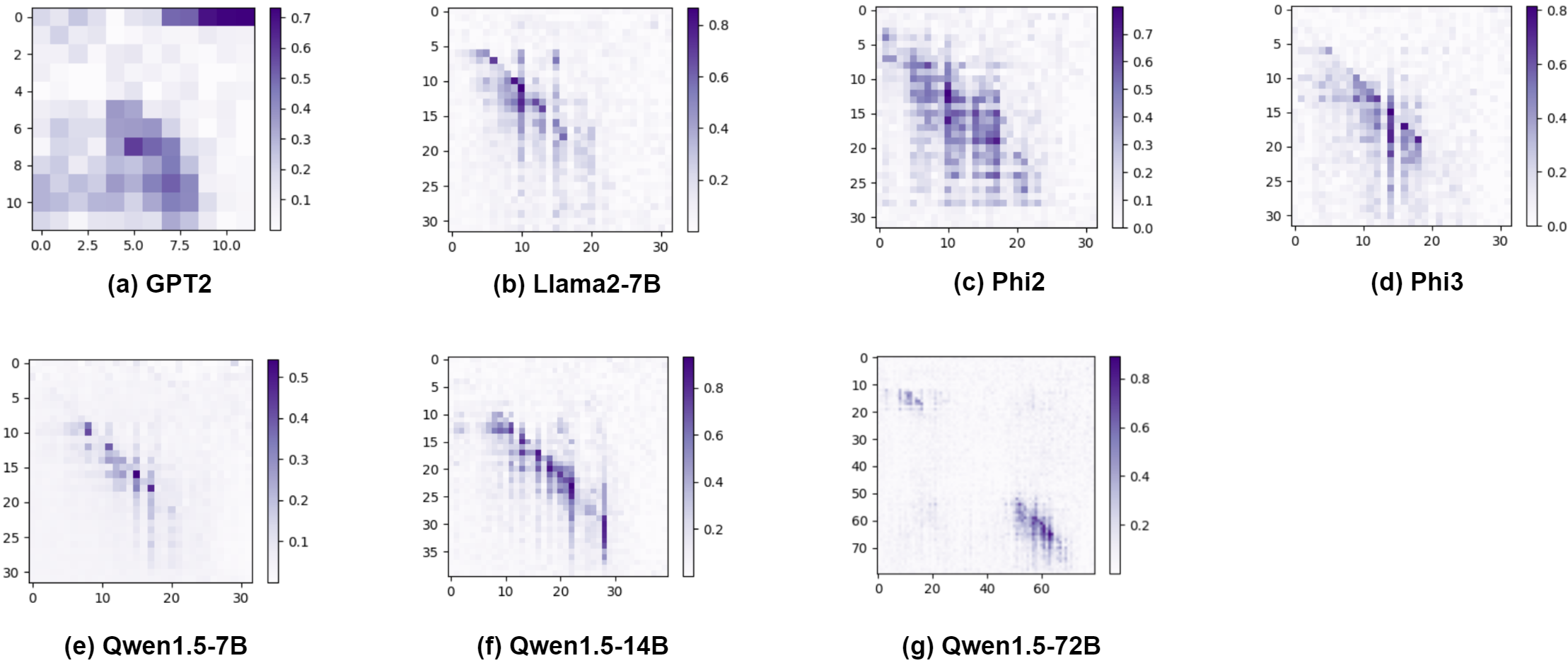}
        \caption{Weight alignment phenomenon for real-world LLMs.}
        \label{fig:other_LLMs}
    \end{figure}

    Except for the phi‑3 large model, the above definitions of the multi‑head matching score and kernel score are equally applicable to other models that do not employ group query attention. The corresponding experimental results are shown in Fig.~\ref{fig:other_LLMs}.

\clearpage

\section{\blue{Causal Intervention}}\label{appendix: causal}

\blue{This section presents causal intervention experiments conducted to verify that the model uses a buffer mechanism when performing symbolic multi-step reasoning tasks. We assume readers are familiar with the content of Section \ref{VTS}. The causal intervention experiments were conducted using a 3-layer Transformer model trained as described in Section \ref{VTS_exp}.}

\begin{figure}[htb]
    \centering
    \includegraphics[width=\textwidth]{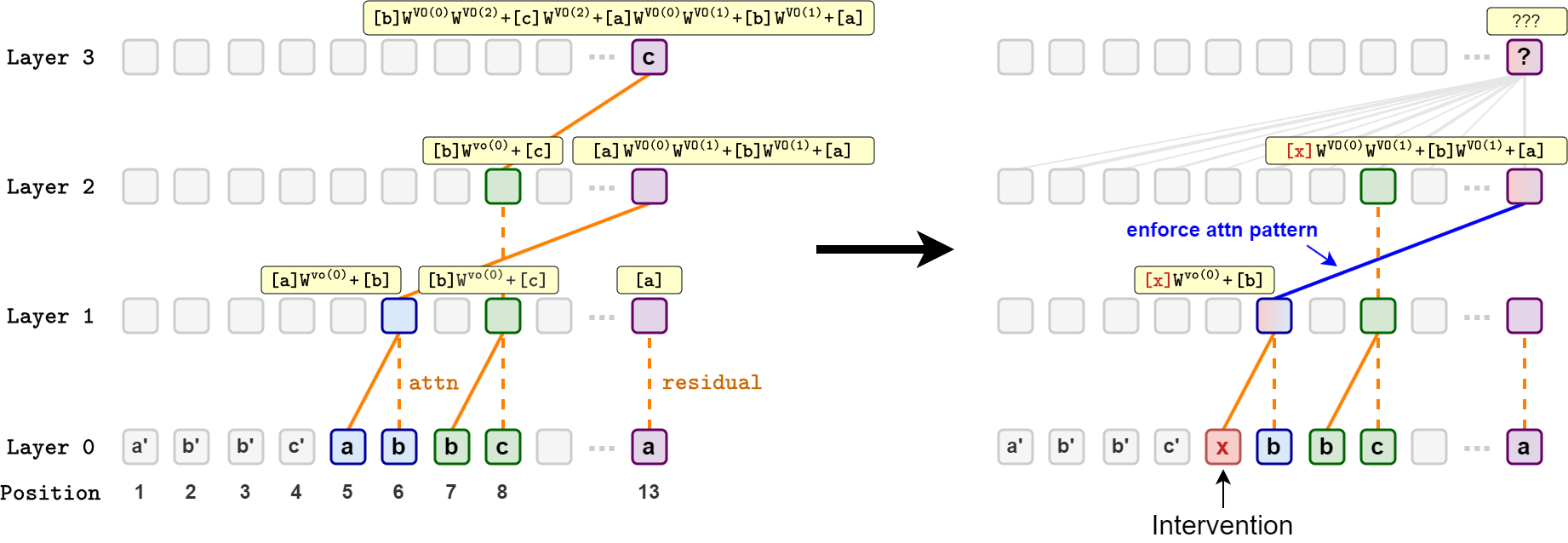}
    \caption{\blue{Logical circuit for 2-step reasoning (left)} \green{and illustration of causal intervention(right). We achieve the goal of intervening in information stored in a specific buffer by modifying the input sequence and enforcing the same attention pattern as before. The figure illustrates how the information $\texttt{[a]}$ stored in the final token's buffer $\mW^{vo(0)}\mW^{vo(1)}$ in layer 2 can be changed to $\texttt{[x]}$.}}\label{fig: 3L1H_specific}
\end{figure}

\blue{First, we identified critical tokens and logical circuits by observing output changes when masking specific attention or residual paths. Fig.~\ref{fig: 3L1H_specific}(left) illustrates the logical circuit of the 3-layer Transformer performing the symbolic 2-step reasoning task. Subsequently, we conducted causal intervention experiments on the information stored in each token.}

\begin{figure}[htb]
    \centering
    \includegraphics[width=1\textwidth]{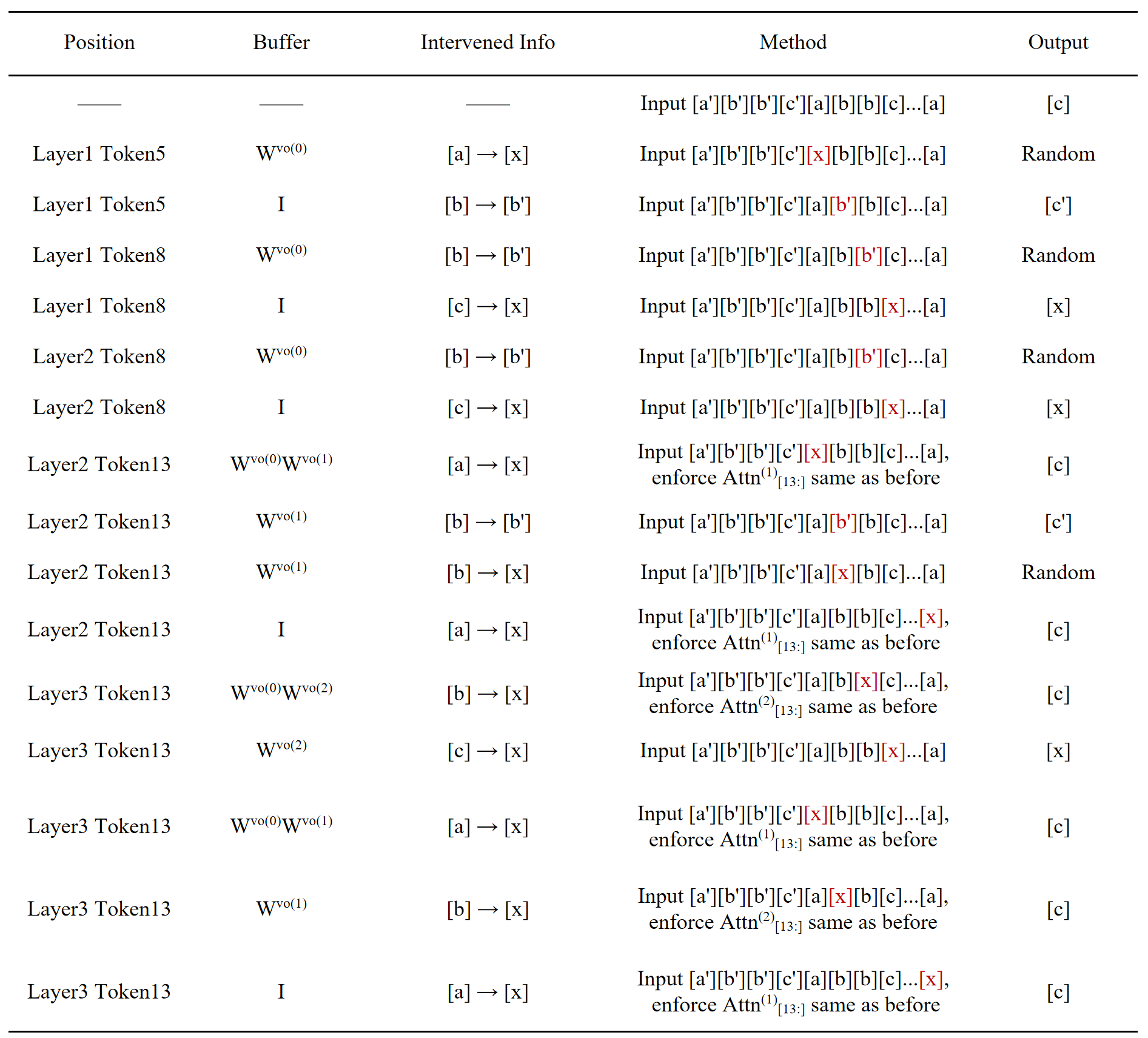}
    \caption{\blue{Causal Intervention Experiment. We performed interventions on the information stored in every buffer of every critical token individually. Here, $\texttt{[x]}$ represents a token that does not appear in the original input. For the final token, only modifying the buffer $\mW^{vo(l-1)}$ in layer $l$ affects the final output.} \green{In this experiment, the tokens in the original sentence are selected from the range [20, 40], while token $\texttt{[x]}$ traverses the range [40, 100]. $\text{Attn}^{(l)}_{[13:]}$ refers to the attention score corresponding to the last token at layer $l$. For the original input, we have $ \text{Attn}^{(1)}_{[13,6]} = 1 $ and $ \text{Attn}^{(2)}_{[13,8]} = 1$. Instances labeled as ``Random" indicate that the output varies erratically as [x] changes. In all other cases, the probability of the model output deviating from the value presented in the table is less than 1e-15.}}\label{fig: causal}
\end{figure}

\blue{Unlike prior works where causal intervention replaced all information within a token with alternative information\citep{feng2023language, meng2022locating, vig2020investigating, wang2024grokked}, we refined the perturbation scope to target a specific buffer within a token. Specifically, we individually replaced the information in each buffer of critical tokens with alternative information to observe the model's output.} \green{For example, as shown in Fig.~\ref{fig: 3L1H_specific}(right), suppose we want to change the information $\texttt{[a]}$ stored in the final token's buffer $\mW^{vo(0)}\mW^{vo(1)}$ in layer 2 to $\texttt{[x]}$. This can be achieved by simply modifying the input sentence $\texttt{[a'][b'][b'][c'][a][b][b][c]...[a]}$ to $\texttt{[a'][b'][b'][c']\textbf{[x]}[b][b][c]...[a]}$, and enforcing $\text{Attn}^{(1)}_{[13:]}$ same as before.} \blue{Fig.~\ref{fig: causal} shows the intervention results for all buffers of all critical tokens.}

\blue{The results reveal that, in layer $l$, the information stored in buffer $W^{vo(l-1)}$ of the final token is crucial. Modifying information in other buffers does not affect the model's output. Combined with the observation in Appendix~\ref{appendix: Matching Operation} that the cosine similarities between $\mW^{vo(0)}, \mW^{vo(1)}, \mW^{vo(2)}, \mI$ are nearly zero, we can confidently assert that the model performs reasoning by leveraging different buffers. This experiment also rules out the possibility of the model only using an overwrite mechanism to perform reasoning.}

\clearpage

\section{Interaction Results with Large Language Models}\label{appendix: QA}

    \begin{figure}[htb]
        \centering
        \includegraphics[width=0.8\textwidth]{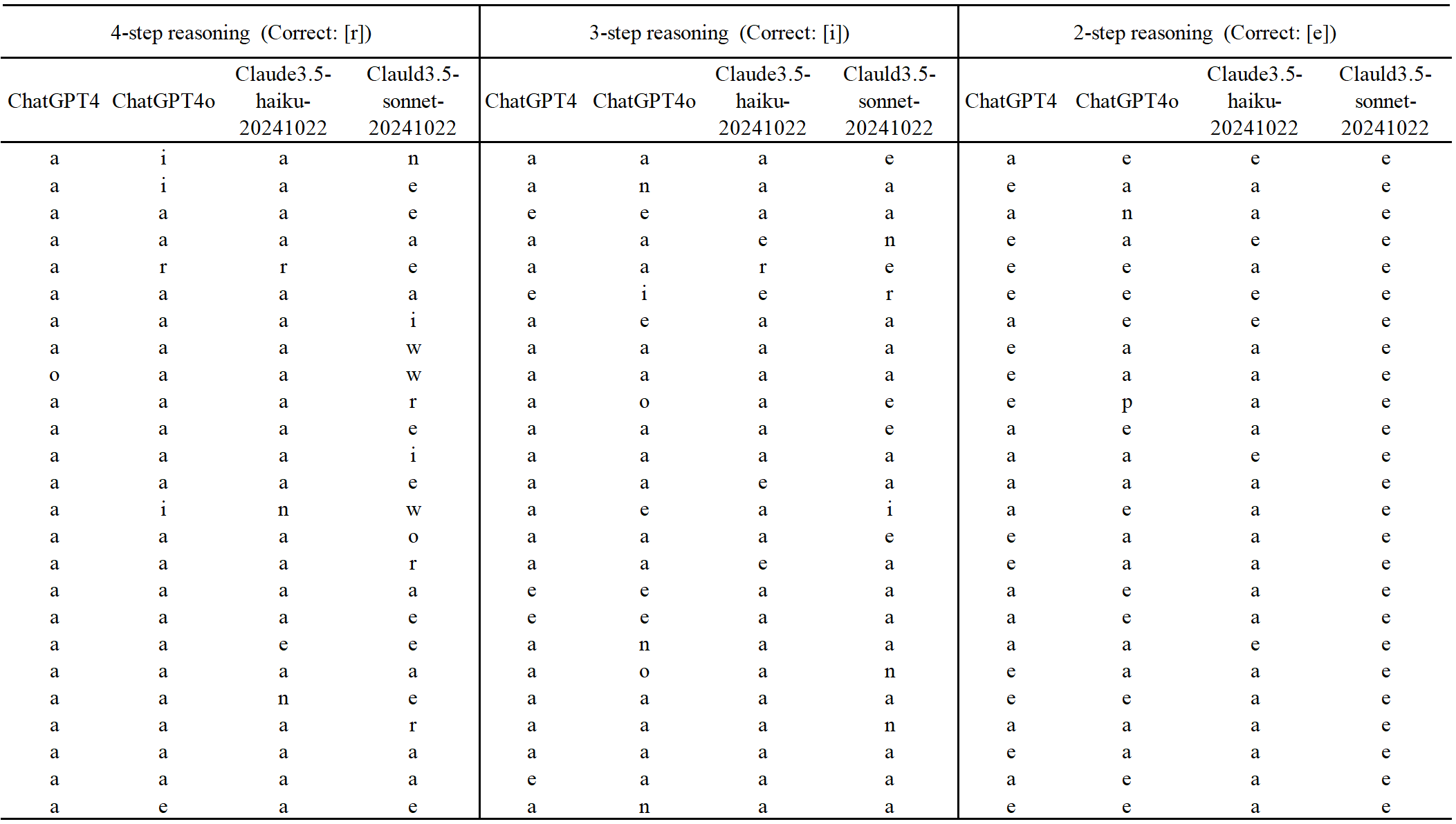}
        \caption{\red{Detailed interaction results with large models. For each type of reasoning task, we tested each large model 25 times. The versions of Claude are Claude 3.5-haiku-20241022 and Claude 3.5-sonnet-20241022.}}
    \end{figure}

\end{document}